\documentclass[runningheads]{llncs}

\usepackage[hidelinks]{hyperref}
\usepackage{amssymb,enumerate,tikz,eqnarray,multirow}
\usetikzlibrary{matrix,arrows,decorations.pathmorphing}
\usepackage{latexsym}
\usepackage{amsfonts,bbding}
\usepackage{bbm}

\def\natnum{{\mathbb N}}

\def\email#1{Email: {\tt #1}}
\def\proof{\noindent{\bf Proof.}\enspace}

\def\qed{~~\hbox{\hskip 1pt \vrule width 4pt height 8pt depth 1.5pt\hskip 1pt}}
\spnewtheorem{thm}{Theorem}{\bf }{\it }
\spnewtheorem{prop}[thm]{Proposition}{\bf }{\it }
\spnewtheorem{prob}[thm]{Open Problem}{\bf }{\it }
\spnewtheorem{cor}[thm]{Corollary}{\bf }{\it }
\spnewtheorem{lem}[thm]{Lemma}{\bf }{\it }
\spnewtheorem{defn}[thm]{Definition}{\bf }{\rm }
\spnewtheorem{rem}[thm]{Remark}{\bf }{\rm }
\spnewtheorem{exmp}[thm]{Example}{\bf }{\rm }
\spnewtheorem{clm}[thm]{Claim}{\bf }{\it }
\spnewtheorem{nota}[thm]{Notation}{\bf }{\rm }
\spnewtheorem{obs}[thm]{Observation}{\bf }{\rm }

\newcommand{\Part}{$\mathit{Part}$}
\newcommand{\ConfPart}{$\mathit{ConfPart}$}

\newcommand{\ConsPart}{$\mathit{ConsPart}$}

\newcommand{\BC}{$\mathit{BC}$}
\newcommand{\Ex}{$\mathit{Ex}$}
\newcommand{\Exf}{$\mathit{Ex^*}$}

\newcommand{\Vac}{$\mathit{Vac}$}
\newcommand{\ConsvEx}{$\mathit{ConsvEx}$}
\newcommand{\BCf}{$\mathit{BC^*}$}
\newcommand{\BCn}{$\mathit{BC^n}$}
\newcommand{\BCa}{$\mathit{BC^a}$}
\newcommand{\BCnPart}{$\mathit{BC^nPart}$}

\newcommand{\Vacf}{$\mathit{Vac^*}$}
\newcommand{\BCfPart}{$\mathit{BC^*Part}$}
\newcommand{\VacfPart}{$\mathit{Vac^*Part}$}
\newcommand{\ConsvPart}{$\mathit{ConsvPart}$}

\newcommand{\Cons}{$\mathit{Cons}$}

\newcommand{\ClsPresv}{$\mathit{ClsPresv}$}

\newcommand{\ApproxPart}{$\mathit{ApproxPart}$}
\newcommand{\WeakApproxPart}{$\mathit{WeakApproxPart}$}
\newcommand{\FinApproxPart}{$\mathit{FinApproxPart}$}
\newcommand{\FinApprox}{$\mathit{FinApprox}$}
\newcommand{\Approx}{$\mathit{Approx}$}
\newcommand{\WeakApprox}{$\mathit{WeakApprox}$}

\newcommand{\FinApproxConsPart}{$\mathit{FinApproxConsPart}$}

\newcommand{\ExfFinApprox}{$\mathit{Ex^*FinApprox}$}
\newcommand{\ClsConsConfPart}{$\mathit{ClsConsConfPart}$}

\newcommand{\FinApproxConfPart}{$\mathit{FinApproxConfPart}$}
\newcommand{\FinApproxConf}{$\mathit{FinApproxConf}$}

\newcommand{\ExKjump}{$\mathit{Ex[K']}$}
\newcommand{\WPart}{$\mathit{WPart}$}

\newcommand{\VacfWPart}{$\mathit{Vac^*WPart}$}

\newcommand{\VacfFinApprox}{$\mathit{Vac^*FinApprox}$}
\newcommand{\FinApproxConsvPart}{$\mathit{FinApproxConsvPart}$}
\newcommand{\ClsPresvFinApprox}{$\mathit{ClsPresvFinApprox}$}

\newcommand{\FinApproxBCf}{$\mathit{FinApproxBC^*}$}

\newcommand{\ConsvPartBC}{$\mathit{ConsvPartBC}$}

\newcommand{\PrudConsvEx}{$\mathit{PrudConsvEx}$}

\newcommand{\FinApproxCons}{$\mathit{FinApproxCons}$}

\newcommand{\ApproxBCfPart}{$\mathit{ApproxBC^*Part}$}
\newcommand{\In}{\mbox{In}}

\newcommand{\RECPart}{$\mathit{RECPart}$}
\newcommand{\RECApproxBCfPart}{$\mathit{RECApproxBC^*Part}$}
\newcommand{\RECAppr}{$\mathit{RECAppr}$}
\newcommand{\oxBCfPart}{$\mathit{oxBC^*Part}$}
\newcommand{\ConsWeakApproxPart}{$\mathit{ConsWeakApproxPart}$}
\newcommand{\ConsApproxPart}{$\mathit{ConsApproxPart}$}
\newcommand{\ConsWeakApprox}{$\mathit{ConsWeakApprox}$}
\newcommand{\ConsWeakApproxBCfPart}{$\mathit{ConsWeakApproxBC^*Part}$}
\newcommand{\ConsApproxBCfPart}{$\mathit{ConsApproxBC^*Part}$}

\begin{document}


\title{Combining Models of Approximation with Partial Learning\thanks{F.~Stephan
was partially supported by NUS grants R146-000-181-112
and R146-000-184-112; S.~Zilles was
partially supported by the Natural Sciences and Engineering Research
Council of Canada (NSERC).}}

\titlerunning{Combining Models of Approximation with Partial Learning}

\author{Ziyuan Gao$^1$, Frank Stephan$^2$ and Sandra Zilles$^3$}

\authorrunning{Z.~Gao, F.~Stephan and S.~Zilles}

\institute{Department of Computer Science\\ University of Regina,
Regina, SK, Canada S4S 0A2\\\email{gao257@cs.uregina.ca}\and
Department of Mathematics and Department of Computer Science\\National
University of Singapore, Singapore
119076\\\email{fstephan@comp.nus.edu.sg} \and Department of Computer
Science\\ University of Regina, Regina, SK, Canada S4S
0A2\\\email{zilles@cs.uregina.ca}}

\maketitle

\begin{abstract}
In Gold's framework of inductive inference, the model of partial
learning requires the learner to output exactly one correct index for
the target object and only the target object infinitely often. Since
infinitely many of the learner's hypotheses may be incorrect, it is
not obvious whether a partial learner can be modified to
``approximate'' the target object.

Fulk and Jain
(Approximate inference and scientific method.
\emph{Information and Computation} 114(2):179--191, 1994)
introduced a model of approximate learning of
recursive functions. The present work extends their
research and solves an open problem of Fulk and Jain by
showing that there is a learner which approximates and
partially identifies every recursive function
by outputting a sequence of hypotheses which, in addition,
are also almost all finite variants of the target function.

The subsequent study is dedicated to the question how these
findings generalise to the learning of r.e.\ languages from
positive data. Here three variants of approximate learning will
be introduced and investigated with respect to the question whether
they can be combined with partial learning.
Following the line of Fulk and Jain's research,
further investigations provide conditions under which partial language
learners can eventually output only finite variants of the target language.
\end{abstract}  

\section{Introduction}

\noindent
Gold \cite{Gold} considered a learning scenario where the
learner is fed with piecewise increasing amounts of finite 
data about a given target language $L$; at every stage where
a new input datum is given, the learner makes a conjecture
about $L$.  If there is exactly one correct
representation of $L$ that the learner always outputs after 
some finite time (assuming that it never stops receiving
data about $L$), then the learner is said to have ``identified  
$L$ in the limit.''  In this paper, it is assumed that
all target languages are encoded as recursively enumerable (r.e.)
sets of natural numbers, and that the learner uses G\"{o}del numbers as its hypotheses.

Gold's learning paradigm has been
used as a basis for a variety of theoretical models in subjects such
as human language acquisition \cite{osh82} and the theory of 
scientific inquiry in the philosophy of science \cite{Cas83,martin98}.  
This paper is mainly concerned with the \emph{partial learning}
model \cite{osh}, which retains several features of Gold's original
framework -- the modelling of learners as recursive functions, 
the use of texts as the mode of data presentation and the 
restriction of target classes to the family of all
r.e.\ sets -- while liberalising the learning criterion
by only requiring the learner to output exactly one hypothesis 
of the target set infinitely often while it must output
any other hypothesis only finitely often. It is known that partial
learning is so powerful that the class of all r.e.\ languages can be
partially learnt~\cite{osh}.

However, the model of partial learning puts no further constraints on
those hypotheses that are output only finitely often. In particular,
it offers no notion of ``eventually being correct'' or even
``approximating'' the target object. From a philosophical point of
view, if partial learning is to be taken seriously
as a model of language acquisition, then it is
quite plausible that learners are capable of
gradually improving the quality of their hypotheses
over time.  For instance, if the learner $M$
sees a sentence $S$ in the text at some
point, then it is conceivable that after 
some finite time, $M$ will only conjecture 
grammars that generate $S$.  This leads
one to consider a notion of the learner 
``approximating'' the target language.

The central question in this paper is whether any partial
learner can be redefined in a way that it approximates the target
object and still partially learns it. The first results,
in the context of partial learning, deal with Fulk and 
Jain's \cite{fulkjain94} notion of approximating recursive 
functions. Fulk and Jain proved the existence of a learner that
``approximates'' every recursive function. This result is
generalised as follows: partial learners can always be made to
approximate recursive functions according to their model and, in
addition, eventually output only finite variants of the target
function, that is, they can be designed as \BCf\ learners\footnote{\BCf\
is mnemonic for ``behaviourally correct with finitely many
anomalies''~\cite{Cas83}.}. This result solves an open question posed
by Fulk and Jain, namely whether recursive functions can be
approximated by \BCf\ learners. Note that \BCf\ learning can also, in
some sense, be considered a form of approximation, as it requires that
eventually all of the hypotheses (including those output only finitely
often) differ from the target object in only finitely many values. It
thus is interesting to see that partial learning can be combined not
only with Fulk and Jain's model of approximation, but also with \BCf\
learning \emph{at the same time}.  Note that in this paper, when two 
learning criteria $A$ and $B$ are said to be combinable, it is generally 
not assumed that the new learner is effectively constructed from the 
$A$-learner and the $B$-learner.

This raises the question whether partial learners can also be turned
into approximate learners in the more general case of learning r.e.\
languages. Unfortunately, Fulk and Jain's model applies only to
learning recursive functions. The second contribution is the design of
three notions of approximate learning of r.e.\ languages, two of which
are directly inspired by Fulk and Jain's model. It is then investigated
under which conditions partial learners can be modified to fulfill the
corresponding constraints of approximate learning. These investigations
are also extended to partial learners with additional constraints,
such as consistency and conservativeness. It will be shown that
while partial learners can always be constructed
in a way so that for any given finite set $D$,
their hypotheses will almost always agree
with the target language on $D$, the same
does not hold if $D$ must be a finite variant
of a fixed infinite set. Thus trade-offs
between certain approximate learning constraints
and partial learning are sometimes unavoidable -- an
observation that perhaps has a broader implication
in the philosophy of language learning.

Following the line of Fulk and Jain's research, conditions are
investigated under which partial language learners can eventually output
only finite variants of the target function. While it remains open
whether or not partial learners for a given \BCf-learnable class
can be made \BCf-learners for this class
without losing identification power, some natural conditions
on a \BCf\ learner $M$ are provided under which all classes learnable by
$M$ can be learnt by some \BCf\ learner $N$ that is at the same time a
partial learner.

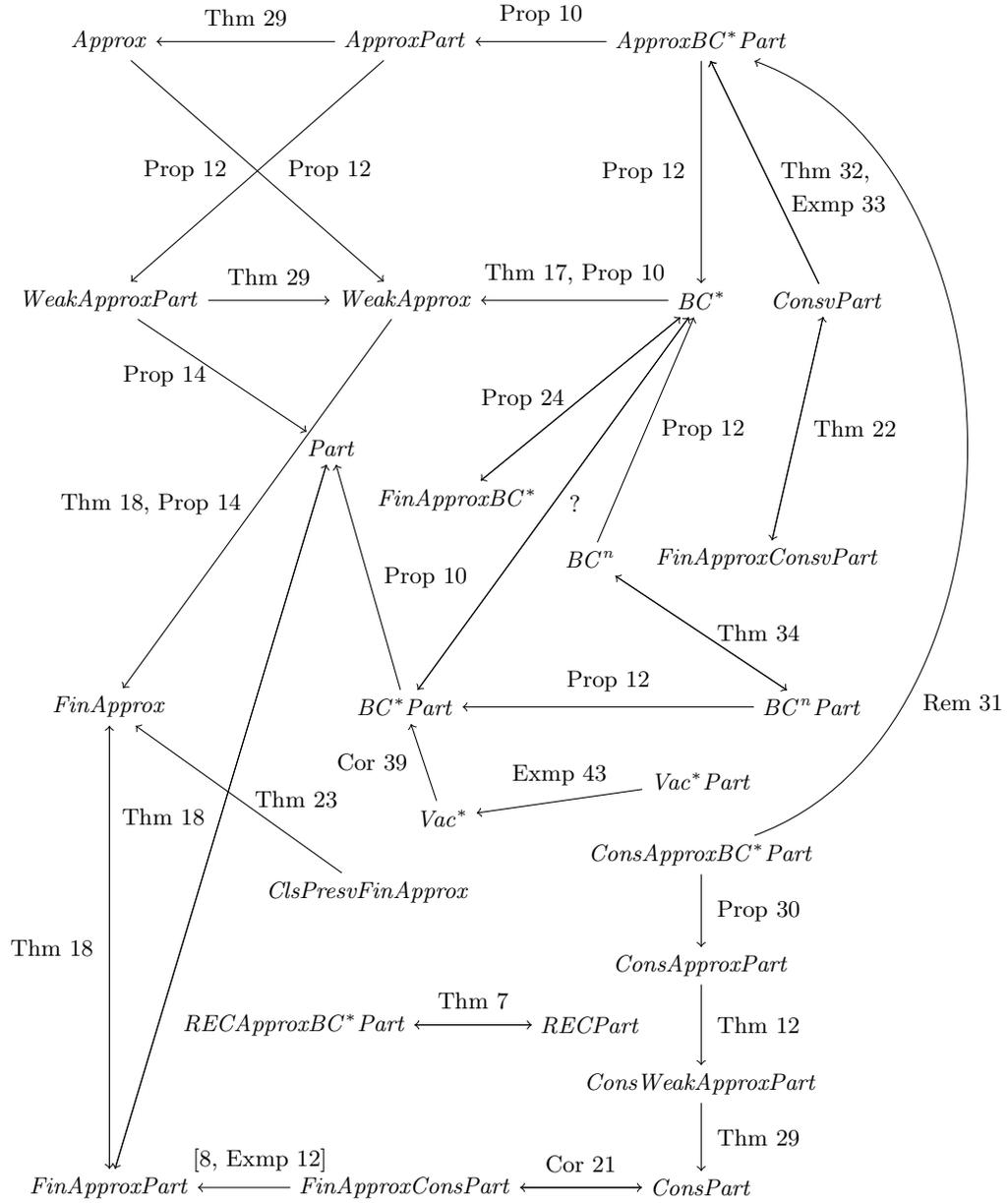
\begin{figure}[!t]\label{ld}
\centering
\begin{tikzpicture}\label{learninghierarchy}[auto]
\node (approx) at (-8,8) {\Approx}; 
\node (weakapprox) at (-4,4.5) {\WeakApprox};
\node (approxpart) at (-4,8) {\ApproxPart};
\node (weakapproxpart) at (-8,4.5) {\WeakApproxPart};
\node (approxbcfpart) at (0,8) {\ApproxBCfPart};
\node (consvpart) at (1.7,4.5) {\ConsvPart};
\node (finapproxconsvpart) at (0.9,1) {\FinApproxConsvPart};
\node (bcf) at (0,4.5) {\BCf};
\node (finapprox) at (-8,-1) {\FinApprox};
\node (finapproxpart) at (-8,-7.5) {\FinApproxPart};
\node (bcn) at (-1.5,1) {\BCn};
\node (bcnpart) at (1.5,-1) {\BCnPart};
\node (finapproxconspart) at (-4,-7.5) {\FinApproxConsPart};
\node (conspart) at (0,-7.5) {\ConsPart};
\node (part) at (-5,2.5) {\Part};
\node (bcfpart) at (-4,-1) {\BCfPart};
\node (recpart) at (-1.5,-5.3) {\RECPart};
\node (recapproxbcfpart) at (-5.5,-5.3) {\RECApproxBCfPart};
\node (finapproxbcf) at (-3.3,1.8) {\FinApproxBCf};
\node (clspresvfinapprox) at (-4.5,-3.5) {\ClsPresvFinApprox};
\node (consweakapproxpart) at (0,-6.1) {\ConsWeakApproxPart};
\node (consapproxpart) at (0,-4.5) {\ConsApproxPart};
\node (vacf) at (-3.5,-2.5) {\Vacf};
\node (vacfpart) at (0,-2) {\VacfPart};
\node (consapproxbcfpart) at (0,-3) {\ConsApproxBCfPart};

\path[->] (approx) edge node[right=0.3cm] {{\footnotesize Prop \ref{bcstarnotappro}}} (weakapprox)
(weakapprox) edge node[left=0.1cm] {{\footnotesize Thm \ref{finapproxpart}, Prop \ref{goldclass}}} (finapprox)
(finapprox) edge node[left=0.1cm] {{\footnotesize Thm \ref{finapproxpart}}} (finapproxpart)
(finapproxpart) edge (finapprox)
(approxpart) edge node[above=0.1cm] {{\footnotesize Thm \ref{th:separation}}} (approx)
(weakapproxpart) edge node[above=0.1cm] {{\footnotesize Thm \ref{th:separation}}} (weakapprox)
(approxpart) edge node[left=0.3cm] {{\footnotesize Prop \ref{bcstarnotappro}}} (weakapproxpart)
(bcf) edge node[above=0.1cm] {{\footnotesize Thm \ref{th:bcstarweakapprox}, Prop \ref{approxinf}}} (weakapprox)
(approxbcfpart) edge node[above=0.1cm] {{\footnotesize Prop \ref{approxinf}}} (approxpart)
(approxbcfpart) edge node[left=0.1cm] {{\footnotesize Prop \ref{bcstarnotappro}}} (bcf)
(consvpart) edge node[right=0.1cm] {{\footnotesize Thm \ref{consvpartimpliesbcfapprox},}} (approxbcfpart)
(consvpart) edge node[right=1cm,below=0.2cm] {{\footnotesize Exmp \ref{sepapproxconsv}}} (approxbcfpart)
(consvpart) edge node[right=0.1cm] {{\footnotesize Thm \ref{finapproxconsvpart}}} (finapproxconsvpart)
(finapproxconsvpart) edge (consvpart)
(weakapproxpart) edge node[left=0.1cm] {{\footnotesize Prop \ref{goldclass}}}(part)
(bcf) edge node[right=0.1cm] {{\footnotesize ?}}(bcfpart)
(bcfpart) edge (bcf)
(bcfpart) edge node[right=0.1cm] {{\footnotesize Prop \ref{approxinf}}}(part)
(bcn) edge node[right=0.1cm] {{\footnotesize Prop \ref{bcstarnotappro}}}(bcf)
(bcnpart) edge node[above=0.1cm] {{\footnotesize Prop \ref{bcstarnotappro}}}(bcfpart)
(bcnpart) edge node[right=0.1cm] {{\footnotesize Thm \ref{thm:bcnpart}}} (bcn)
(bcn) edge (bcnpart)
(finapproxconspart) edge node[above=0.1cm] {{\footnotesize ~~~~\cite[Exmp 12]{gaostezill13}}}(finapproxpart)
(conspart) edge node[above=0.1cm] {{\footnotesize Cor \ref{finapproxconspart}}}(finapproxconspart)
(finapproxconspart) edge (conspart)
(finapproxpart) edge node[left=0.1cm] {{\footnotesize Thm \ref{finapproxpart}}}(part)
(part) edge (finapproxpart)
(recpart) edge node[above=0.1cm] {{\footnotesize Thm \ref{recbcpartappro}}}(recapproxbcfpart)
(recapproxbcfpart) edge (recpart)
(finapproxbcf) edge node[left=0.1cm] {{\footnotesize Prop \ref{finapproxbcf}}} (bcf)
(bcf) edge (finapproxbcf)
(clspresvfinapprox) edge node[right=0.1cm] {{\footnotesize Thm \ref{approxnotclspresv}}}(finapprox)
(consweakapproxpart) edge node[right=0.1cm] {{\footnotesize Thm \ref{th:separation}}} (conspart)
(vacf) edge node[left=0.1cm] {{\footnotesize Cor \ref{vacstarimpbcpart}}} (bcfpart)
(vacfpart) edge node[above=0.1cm] {{\footnotesize Exmp \ref{sepvacfpart}}} (vacf)
(consapproxpart) edge node[right=0.1cm] {{\footnotesize Thm \ref{bcstarnotappro}}} (consweakapproxpart)
(consapproxbcfpart) edge node[right=0.1cm] {{\footnotesize Prop \ref{propsep:consapproxpartnotbcf}}} (consapproxpart);
\draw[bend right=70,->]  (consapproxbcfpart) to node [auto, near start, swap] {\footnotesize Rem \ref{prop:sepapproxbcfpartcons}} (approxbcfpart);
\end{tikzpicture}
\caption{Learning hierarchy} \label{fig:1}
\end{figure}

Figure \ref{fig:1} summarises the main results of this paper.
\RECPart\ and \RECAppr- \oxBCfPart\ refer respectively to partial
learning of recursive functions and
approximate \BCf\ partial learning of recursive functions.
The remaining learning criteria  
are abbreviated (see Definitions \ref{defnmain}, \ref{approxfunctiondef} and \ref{approxdef}), 
and denote learning of classes of r.e.\ languages.
An arrow from 
criterion $A$ to criterion $B$ means that the collection of
classes learnable under model $A$ is contained in that
learnable under model $B$.  Each arrow is labelled with
the Corollary/Example/Remark/Theorem number(s) that 
proves (prove) the relationship represented by the arrow.  
If there is no path from $A$ to $B$, then
the collection of classes learnable under model $A$ is not contained in that
learnable under model $B$.

\section{Preliminaries}

\noindent
The notation and terminology from recursion theory adopted in this
paper follows in general the book of Rogers \cite{Rog}.  Background
on inductive inference can be found in \cite{jainsys}.
The symbol $\mathbb{N}$ denotes
the set of natural numbers, $\{0,1,2,\ldots\}$.
Let $\varphi_0,\varphi_1, \varphi_2,\ldots$ denote a fixed acceptable
numbering \cite{Rog} of all partial-recursive functions over $\mathbb N$.
Given a set $S$, 
$S^*$ denotes the set of all finite sequences in $S$.
Wherever no confusion may arise, $S$
will also denote its own characteristic function,
that is, for all $x\in\natnum$, $S(x) = 1$ if $x \in S$
and $S(x) = 0$ otherwise.
One defines the $e$-th r.e.\ set $W_e$ as $dom(\varphi_e)$
and the $e$-th canonical finite set by choosing $D_e$
such that $\sum_{x\in D_e} 2^x = e$.
This paper fixes a one-one padding function $pad$
with $W_{pad(e,d)} = W_e$ for all $e,d$.
Furthermore, $\langle x,y\rangle$ denotes Cantor's pairing function,
given by $\langle x,y\rangle =\frac{1}{2}(x+y)(x+y+1)+y$.
A triple $\langle x,y,z\rangle$ denotes $\langle \langle x,y\rangle,z\rangle$.
The notation $\eta(x)\!\downarrow$ means that $\eta(x)$ is defined, and
$\eta(x)\!\uparrow$ means that $\eta(x)$ is undefined.
The notation
$\varphi_e(x)\!\uparrow$ means that $\varphi_e(x)$ remains undefined and
$\varphi_{e,s}(x)\!\downarrow$ means that $\varphi_e(x)$ is defined within
$s$ steps, that is, the computation of $\varphi_e(x)$ halts within $s$ steps.
$K$ denotes the halting problem, that is, $K = \{x: \varphi_x(x)\downarrow\}$.
For any r.e.\ set $A$, $A_s$ denotes the $s$th approximation
of $A$; it is assumed that for all $s$, $|A_{s+1} - A_s| \leq 1$
and $A_s \subseteq \{0,\ldots,s\}$.

For any $\sigma,\tau\in (\mathbb{N}\cup \{\#\})^*, \sigma \preceq
\tau$ if and only if $\sigma$ is a prefix of $\tau$,
$\sigma \prec \tau$ if and only if $\sigma$ is a proper prefix of
$\tau$, and $\sigma(n)$ denotes the element in the $n$th position of
$\sigma$, starting from $n=0$.  For each $\sigma \neq \epsilon$, 
$\sigma'$ denotes the string obtained from $\sigma$ by
deleting the last symbol of $\sigma$.
The concatenation of two strings $\sigma$ and $\tau$ shall be
denoted by $\sigma\circ\tau$; for convenience, and whenever there is no
possibility of confusion, this is occasionally
denoted by $\sigma\tau$.
Let $\sigma[n]$ denote the
sequence $\sigma(0)\circ\sigma(1)\circ\ldots\circ\sigma(n-1)$.
The length of $\sigma$ is denoted by $|\sigma|$.

\section{Learning}

The basic learning paradigms studied in the present paper
are 
\emph{behaviourally
correct learning} \cite{Barz74,Cas82} and \emph{partial learning}
\cite{osh}.  These learning models assume that the learner is 
presented with just positive examples of the target language, 
and that the learner is fed with a finite 
amount of data at every stage. They are modifications of the model of
explanatory learning (or
``learning in the limit''), first introduced by Gold \cite{Gold}, 
in which the learner must output in the limit 
a single correct representation $h$ of the target language $L$;
if $L$ is an r.e.\ set, then $h$ is usually an
r.e.\ index of $L$ with respect to the standard
numbering $W_0,W_1,W_2,\ldots$ of all r.e.\ sets.
B\={a}rzdi\c{n}\u{s} \cite{Barz74} and Case \cite{Cas82}
considered the more powerful model of behaviourally correct learning, whereby
the learner must almost always output a correct hypothesis
of the input set, but some of the correct hypotheses may be syntactically
distinct.  Case and Smith \cite{Cas83} also introduced
a less stringent variant of \BC\ learning of recursive functions, 
\BCf\ learning, which only requires the learner to output
in the limit finite variants of the target recursive function.  
Still more general is the criterion of partial learning that Osherson, 
Stob and Weinstein \cite{osh} defined; in this model, the 
learner must output exactly one correct
index of the input set infinitely often and output any
other conjecture only finitely often.  

One can also impose constraints on the quality of a learner's 
hypotheses.  For example, Angluin \cite{Ang80} introduced
the notion of \emph{consistency}, which is the requirement
that the learner's hypotheses must enumerate at least 
all the data seen up to the current stage.  This seems
to be a fairly natural demand on the learner, for
it only requires that the learner's conjectures
never contradict the available data on the target
language.  Angluin \cite{Ang80} also introduced
the learning constraint of \emph{conservativeness};
intuitively, a conservative learner never makes
a mind change unless its prior conjecture does
not enumerate all the current data.  
A further constraint proposed by Osherson, Stob and Weinstein
\cite{osh82} is \emph{confidence}, according to
which the learner must converge on any (even non
r.e.) text.  These three learning criteria have since 
been adapted to the partial learning model 
\cite{consvpart,gaostezill13}.

Lange and Zeugmann \cite{lange} showed that learning in the
limit is less powerful if the hypothesis space of the
learner is restricted to the target class.  It would
thus be quite natural to ask whether this constraint
on the hypothesis space of the learner has a similar
effect on partial learning or on approximate learning.
For this purpose, one distinguishes between \emph{class-comprising}
learning and \emph{class-preserving} learning \cite{lange}.
If the learner $M$ only conjectures languages that
it can successfully learn, then $M$ is said to be
\emph{prudent} \cite{osh82}.  The learning criteria 
discussed so far (and, where applicable, their partial 
learning analogues) are formally introduced below.

\begin{defn}\cite{ZLK95}
$M$ is said to \emph{class-comprisingly}
learn $\mathcal{C}$ if it learns $\mathcal{C}$
with respect to a hypothesis space $\{H_0,H_1,H_2,\ldots\}$,
where $H_0,H_1,H_2,\ldots$ are r.e.\ sets,
such that $\mathcal{C} \subseteq \{H_0,H_1,H_2,\ldots\}$.
\end{defn}
        
\begin{defn}\cite{ZLK95}
$M$ is said to \emph{class-preservingly} (\ClsPresv)
learn $\mathcal{C}$ if it learns $\mathcal{C}$
with respect to a hypothesis space $\{H_0,H_1,H_2,\ldots\}$,
where $H_0,H_1,H_2,\ldots$ are r.e.\ sets,
such that $\mathcal{C} = \{H_0,H_1,H_2,\ldots\}$.
\end{defn}

\noindent Throughout this paper, 
successful learning with respect to a class $\mathcal{C}$ 
will generally refer to class-comprising learning unless specified
otherwise. 

The learning criteria 
discussed so far (and, where applicable, their partial 
learning analogues) are formally introduced below.

Let ${\mathcal{C}}$ be a class of r.e.\ sets.
Throughout this paper, the mode of data presentation is
that of a \emph{text}, by which is meant an
infinite sequence of natural numbers and the \# symbol. 
Formally, a \emph{text} $T_L$ for some
$L$ in $\mathcal{C}$
is a map $T_L:\mathbb{N} \rightarrow \mathbb{N}
\cup \{\#\}$
such that $L = range(T_L)$;
here, $T_L[n]$ denotes the sequence $T_L(0)\circ
T_L(1)\circ\ldots\circ T_L(n-1)$ and the
range of a text $T$, denoted $range(T)$,
is the set of numbers occurring in $T$. 
Analogously, for a finite sequence $\sigma$,
$range(\sigma)$ is the set of numbers
occurring in $\sigma$.
A text, in other words, is a presentation
of positive data from the target set.
A \emph{learner}, denoted by $M$ in the
following definitions, is a recursive
function mapping $(\mathbb{N}\cup\{\#\})^*$ into
$\mathbb{N}$.
$M$ may also be equipped with an oracle. In this case,
a learner that has access to oracle $A$ is an 
$A$-recursive function mapping $(\mathbb{N}\cup\{\#\})^*$ 
into $\mathbb{N}$. 

\begin{defn}\label{defnmain}
\begin{enumerate}[{\sc (i)}]
\item \cite{osh} $M$ \emph{partially} (\Part) learns $\mathcal{C}$
if, for every $L$ in $\mathcal{C}$ and each text $T_L$
for $L$, there is exactly one index $e$ such that
$M(T_L[k]) = e$ for infinitely many $k$; furthermore, if $M$ outputs
$e$ infinitely often on $T_L$, then $L = W_e$.

\item \cite{Cas82} $M$ \emph{behaviourally correctly} (\BC)
learns $\mathcal{C}$ if, for every $L$ in $\mathcal{C}$
and each text $T_L$ for $L$, there is a number $n$ for which
$L = W_{M(T_L[j])}$ whenever $j\geq n$.

\item \cite{Ang80} $M$ is \emph{consistent} (\Cons)
if for all $\sigma \in (\mathbb{N} \cup \{\#\})^*$,
$range(\sigma) \subseteq W_{M(\sigma)}$.

\item \cite{Ang80} For any text $T$, $M$ is \emph{consistent on $T$} if
$range(T[n]) \subseteq W_{M(T[n])}$
for all $n > 0$.

\item \cite{gaostezill13} $M$ is said to \emph{consistently 
partially} (\ConsPart) learn $\mathcal{C}$ if it partially 
learns $\mathcal{C}$ from text and is consistent.

\item \cite{consvpart} $M$ is said to \emph{conservatively
partially} (\ConsvPart) learn
$\mathcal{C}$ if it partially
learns $\mathcal{C}$ and outputs on each
text for every $L$ in $\mathcal{C}$ exactly one index
$e$ with $L \subseteq W_e$.

\item \cite{gaostezill13} $M$ is said to
\emph{confidently partially} (\ConfPart) learn
$\mathcal{C}$ if it partially learns $\mathcal{C}$ from
text and outputs on every infinite sequence
(including sequences that are not texts for any member of $\mathcal{C}$) 
exactly one index infinitely often.

\item \cite{Cas83} $M$ is said to \emph{behaviourally correctly} 
learn $\mathcal{C}$ \emph{with at most $a$ anomalies} (\BCa) iff 
for every $L \in \mathcal{C}$ and each text $T_L$ for $L$, 
there is a number $n$ for which $|(W_{M(T_L[j])}-L)\cup(L-W_{M(T_L[j])})|\leq a$
whenever $j \geq n$.

\item \cite{Cas83} $M$ is said to \emph{behaviourally correctly} 
learn $\mathcal{C}$ \emph{with finitely many anomalies} (\BCf) iff for
every $L \in \mathcal{C}$
and each text $T_L$ for $L$, there is a number
$n$ for which $|(W_{M(T_L[j])}-L)\cup(L-W_{M(T_L[j])})|<\infty$
whenever $j \geq n$.

\end{enumerate}
\end{defn}

\noindent
This paper will also consider combinations of different
learning criteria; for learning criteria $A_1,\ldots,A_n$,
a class $\mathcal{C}$ is said to be $A_1\ldots A_n$-learnable
iff there is a learner $M$ such that $M$
$A_i$-learns $\mathcal{C}$ for all $i\in\{1,\ldots,n\}$.  

\section{Approximate Learning of Functions}

Fulk and Jain \cite{fulkjain94} proposed
a mathematically rigorous definition of 
\emph{approximate inference}, a notion 
originally motivated by studies in the 
philosophy of science. 

\begin{defn}\cite{fulkjain94}\label{approxfunctiondef}
An approximate (\Approx) learner outputs on the graph of a function $f$ a
sequence of hypotheses such that there is a sequence $S_0,S_1,\ldots$
of sets satisfying the following conditions:

\noindent(a) The $S_n$ form an ascending sequence of sets such that
their union is the set of all natural numbers;

\noindent(b) There are infinitely many $n$ such that $S_{n+1}-S_n$ is infinite;

\noindent(c) The $n$-th hypothesis is correct on all $x \in S_n$ but
nothing is said about the $x \notin S_n$.
\end{defn}

\noindent
The next proposition simplifies this set of conditions. 

\begin{prop}\label{approcond}
$M$ \Approx\ learns a recursive function $f$ iff
the following conditions hold:

\noindent(d) For all $x$ and almost all $n$, $M$'s $n$-th hypothesis
is correct at $x$;

\noindent(e) There is an infinite set $S$ such that for almost all $n$
and all $x \in S$, $M$'s $n$-th hypothesis is correct at $x$.
\end{prop}

\proof
If one has (a), (b), (c), then the set $S$ is just the first set $S_n$
which is infinite and the other conditions follow.

If one has (d) and (e) and one distinguishes two cases: If $n$ is so
small that the $n$-th and all subsequent hypotheses are not yet
correct on $S$ then one lets $S_n = \emptyset$ else one defines that
$S_n$ contains all $x \leq n$ such that each $m$-th hypothesis with $m
\geq n$ is correct on $x$ plus half of those members of $S$ which are
not in any $S_m$ with $m<n$. So the trick is just not to put all
members of $S$ at one
step into some $S_n$ but just to put at each step which is applicable
an infinite new amount while still another infinite amount remains
outside $S_n$ to be put later.~\qed 

\medskip
\noindent
Fulk and Jain interpreted their notion of approximation
as a process in scientific inference whereby physicists
take the limit of the average result of a sequence of 
experiments.  Their result that the class of recursive
functions is approximately learnable seems to justify
this view.

\begin{thm}[Fulk and Jain \cite{fulkjain94}]\label{recfuncapprox}
There is a learner $M$ that \Approx\ learns every
recursive function. 
\end{thm}

\noindent The following theorem answers an 
open question posed by Fulk and Jain \cite{fulkjain94} on
whether the class of recursive functions has a learner
which outputs a sequence of hypotheses that approximates
the function to be learnt and almost always differs from the
target only on finitely many places.

\begin{thm}\label{recbcpartappro}
There is a learner $M$ which learns the class of all recursive
functions such that (i) $M$ is a \BCf\ learner, (ii) $M$ is
a partial learner and (iii) $M$ is an approximate learner.
\end{thm}

\proof
Let $\psi_0,\psi_1,\ldots$ be an enumeration of all recursive
functions and some partial ones such that in every step $s$ there is
exactly one pair $(e,x)$ for which $\psi_e(x)$ becomes defined at step
$s$ and this pair satisfies in addition that $\psi_e(y)$ is already
defined by step $s$ for all $y < x$.
Furthermore, a function $\psi_e$ is said to make progress on
$\sigma$ at step $s$ iff $\psi_e(x)$ becomes defined at step $s$
and $x \in dom(\sigma)$ and $\psi_e(y) = \sigma(y)$ for all $y \leq x$.

Now one defines for every $\sigma$ a partial-recursive
function $\vartheta_{e,\sigma}$ as follows:
\begin{itemize}
\item $\vartheta_{e,\sigma}(x) = \sigma(x)$ for all $x \in dom(\sigma)$;
\item Let $e_t = e$;
\item Inductively for all $s \geq t$, if some index $d < e_s$
      makes progress on $\sigma$ at step $s+1$ then let $e_{s+1} = d$
      else let $e_{s+1} = e_s$;
\item For each value $x \notin dom(\sigma)$,
      if there is a step $s \geq t+x$ for which $\psi_{e_s,s}(x)$ is
      defined then $\vartheta_{e,\sigma}(x)$ takes this value for the
      least such step $s$, else $\vartheta_{e,\sigma}(x)$ remains
      undefined.
\end{itemize}
The learner $M$, now to be constructed, uses these functions
as hypothesis space; on input $\tau$, $M$ outputs the index of
$\vartheta_{e,\sigma}$ for the unique $e$ and shortest prefix $\sigma$
of $\tau$ such that the following three conditions are satisfied at
some time $t$:
\begin{itemize}
\item $t$ is the first time such that $t \geq |\tau|$ and some function
      makes progress on $\tau$;
\item $\psi_e$ is that function which makes progress at $\tau$;
\item for every $d<e$, $\psi_d$ did not make progress on $\tau$
      at any $s \in \{|\sigma|,\ldots,t\}$ and either
      $\psi_{d,|\sigma|}$ is inconsistent with $\sigma$
      or $\psi_{d,|\sigma|}(x)$ is undefined for at least one
      $x \in dom(\sigma)$.
\end{itemize}
For finitely many strings $\tau$ there might not be any such
function $\vartheta_{e,\sigma}$, as 
$\tau$ is required to be longer than the largest value up to which some
function has made progress at time $|\tau|$, which can be guaranteed only 
for almost all $\tau$. For these finitely
many exceptions, $M$ outputs a default hypothesis, e.g., for
the everywhere undefined function. Now the three conditions
(i), (ii) and (iii) of $M$ are verified. For this, let $\psi_d$
be the function to be learnt, note that $\psi_d$ is total.

Condition (i): $M$ is a \BCf\ learner. Let $d$ be the least index of
the function $\psi_d$ to be learnt and let $u$ be the last step where
some $\psi_e$ with $e < d$ makes progress on $\psi_d$.  Then every
$\tau \preceq \psi_d$ with $|\tau| \geq u+1$ satisfies that first
$M(\tau)$ conjectures a function $\vartheta_{e,\sigma}$ with $e \geq
d$ and $|\sigma| \geq u+1$ and $\sigma \preceq \psi_d$ and second that
almost all $e_s$ used in the definition of $\vartheta_{e,\sigma}$ are
equal to $d$; thus the function computed is a finite variant of
$\psi_d$ and $M$ is a \BCf\ learner.

Condition (ii): $M$ is a partial learner. Let $t_0,t_1,\ldots$
be the list of all times where $\psi_d$ makes progress on itself
with $u < t_0 < t_1 < \ldots$. Note that whenever $\tau \preceq
\psi_d$ and $|\tau| = t_k$ for some $k$ then the conjecture
$\vartheta_{e,\sigma}$ made by $M(\tau)$ satisfies $e = d$ and
$|\sigma| = u+1$. As none of these conjectures make progress from step $u+1$
onwards on $\psi_d$, they also do not make progress on $\sigma$ after
step $|\sigma|$ and $\vartheta_{e,\sigma} = \psi_d$; hence the learner
outputs some index for $\psi_d$ infinitely often. Furthermore,
all other indices $\vartheta_{e,\sigma}$ are output only finitely
often: if $e < d$ then $\psi_e$ makes no progress
on the target function $\psi_d$ after step $u$; if $e > d$ then
the length of $\sigma$ depends on the prior progress of $\psi_d$
on itself, and if $|\tau| > t_k$ then $|\sigma| > t_k$.

Condition (iii): $M$ is an approximate learner.  Conditions (d) and (e)
in Proposition \ref{approcond} are used. Now it is shown that, for all $\tau
\preceq \psi_d$
with $t_k \leq |\tau| < t_{k+1}$, the hypothesis
$\vartheta_{e,\sigma}$ issued by $M(\tau)$ is
correct on the set $\{t_0,t_1,\ldots\}$.  If $|\tau| = t_k$ then the
hypothesis is correct everywhere as shown under condition (ii). So
assume that $e>d$. Then $|\tau| > t_k$
and $|\sigma| > t_k$, hence $\vartheta_{e,\sigma}(x) = \psi_d(x)$ for
all $x \leq t_k$.  Furthermore, as $\psi_d$ makes progress on $\sigma$
in step $t_{k+1}$ and as no $\psi_c$ with $c<d$ makes progress on
$\sigma$ beyond step $|\sigma|$, it follows that the $e_s$ defined in
the algorithm of $\vartheta_{e,\sigma}$ all satisfy $e_s = d$ for $s
\geq t_{k+1}$; hence $\vartheta_{e,\sigma}(x) = \psi_d(x)$ for all $x
\geq t_{k+1}$.~\qed

\section{Approximate Learning of Languages}\label{languagelearn}

This section proposes three notions of approximation in
language learning.  The first two notions, \emph{approximate}
learning and \emph{weak approximate} learning, are adaptations
of the set of conditions for approximately learning recursive
functions given in Proposition \ref{approcond}.
Recall that a set $V$ is a finite variant of 
a set $W$ iff there is an $x$ such that for
all $y>x$ it holds that $V(y) = W(y)$.

\begin{defn}\label{approxdef}
Let $S$ be a class of languages. $S$ is \emph{approximately (\Approx)
learnable} iff there is a learner $M$ such that
for every language $L \in S$ there is an infinite set
$W$ such that for all texts $T$ and
all finite variants $V$ of $W$ and almost
all hypotheses $H$ of $M$ on $T$, $H \cap V = L \cap V$.
$S$ is \emph{weakly approximately (\WeakApprox) learnable} iff
there is a learner $M$ such that for every language $L \in S$ and for every text $T$ for $L$
there is an infinite set $W$ such that for all
finite variants $V$ of $W$ and almost
all hypotheses $H$ of $M$ on $T$, $H \cap V = L \cap V$. $S$ 
is \emph{finitely approximately 
(\FinApprox) learnable} iff there is a learner $M$ such that 
for every language $L \in S$, 
all texts $T$ for $L$, and any finite set $D$, it holds 
that for almost all hypotheses $H$ of $M$ on $T$, 
$H \cap D = L \cap D$.
\end{defn}

\begin{rem}
Jain, Martin and Stephan \cite{jms13} defined a partial-recursive
function $C$ to be an \emph{\In -classifier} for a class
$S$ of languages if, roughly speaking, for
every $L\in S$, every text $T$ for $L$,
every finite set $D$, and almost all $n$,
$C$ on $T[n]$ will correctly ``classify" all
$x\in D$ as either belonging to $L$ or not
belonging to $L$. 
A learner $M$ that \FinApprox\ learns a class $S$
may be translated into a total \In -classifier for
$S$, and vice versa.  
\end{rem}

\noindent
Approximate learning requires, for each target language, the existence
of a set $W$ suitable for all texts, while in weakly
approximate learning the set $W$ may depend on 
$T$.  In the weakest notion, finitely approximate learning, on 
any text $T$ for a target language $L$ the learner is 
only required to be almost always correct on any
finite set.  
As will be seen later, this model is so powerful that the whole
class of r.e.\ sets can be finitely approximated
by a partial learner.  The following results illustrate the models of
approximate and weakly approximate learning.
They establish that, in contrast to the
function learning case, approximate language learnability does not
imply \BCf\ learnability. \BCf\ learnability does not imply
approximate learnability either, but weakly approximate learning is
powerful enough to cover all \BCf\ learnable classes.

\begin{prop}\label{approxinf}
If there is an infinite r.e.\ set $W$ such that all members
of the class contain $W$ then the class is \Approx\ learnable.
\end{prop}

\proof
The learner for this just conjectures $range(\sigma) \cup W$ on
any input $\sigma$.~\qed

\medskip 
\noindent
Thus approximate learning does, for languages, not imply \BCf\ learning.
\footnote{For example, take the class of all supersets
of the set of even numbers.}
Note that for infinite coinfinite r.e.\ sets $W$, the class of all
r.e.\ supersets of $W$ is not \BCf\ learnable.
The next result is the mirror image of the previous result by just
considering a learner which conjectures the range of the data
seen so far; for each set $L$ in the class the infinite set $S$
in item (e) of Proposition~\ref{approcond} is just the complement of $L$.  
 
\begin{prop}\label{approxcoinfinite}
If a class $\mathcal{C}$ consists only of coinfinite r.e.\ sets
then $\mathcal{C}$ is \Approx\ learnable.
\end{prop}

\medskip
\noindent
While the class of all coinfinite r.e.\ sets can be approximated,
this is not true for the class of all cofinite sets.

\begin{prop}\label{bcstarnotappro}
The class of all cofinite sets is \ConsWeakApproxBCfPart\ learnable but neither
\Approx\ learnable nor \BCn\ learnable for any $n$.
\end{prop}

\proof
To make a \ConsWeakApproxBCfPart\ learner,
define $P$ as follows.  On input $\sigma$, $P$ determines
whether or not $range(\sigma)-range(\sigma') = \{x\}$
for some $x \in \natnum$.
If $range(\sigma)-range(\sigma')$ is either empty
or equal to $\{\#\}$, then $P$ repeats its last conjecture 
($P(\sigma')$) if $\sigma' \neq \epsilon$; if $\sigma' = \epsilon$,
then $P$ outputs a default hypothesis, say a canonical
index for $\natnum$.
If $range(\sigma)-range(\sigma') = \{x\}$ for some
$x\in\natnum$, then $P$ determines the maximum
$w$ (if such a $w$ exists) such that $w \notin range(\sigma) 
\cap \{0,\ldots,x\}$, and outputs a canonical
index for the cofinite set $(range(\sigma) \cap
\{0,\ldots,w\}) \cup \{z: z > w\}$.
If no such $w$ exists, then $P$ outputs a canonical
index for $\natnum$. 

Given any text $T$ for a cofinite set $L \neq \natnum$
such that $w = \max(\natnum-L)$, there is a sufficiently
large $s$ such that $range(T[s'+1]) \cap \{0,\ldots,w\}
= L \cap \{0,\ldots,w\}$ for all $s' > s$.
Furthermore, there are infinitely many $n > s$ such that
$range(T[n+1]) - range(T[n]) = \{x\}$ for
some number $x > w$, and on each of these text
prefixes $T[n+1]$, $P$ will output
a canonical index for $L$.  $P$ is also consistent
by construction.  Thus $P$ consistently partially learns
$L$. On any text $T'$ for $\natnum$, there are infinitely
many stages $n$ at which $range(T'[n+1])$ contains
all numbers less than $x$ for some $x$, and therefore
$P$ will output a canonical index for $\natnum$
infinitely often.  To see that $P$ is also
a \WeakApprox\ learner, observe that 
if $T''$ is a text for a cofinite set $L$,
then $T''$ contains an infinite subsequence
$T''(n_0),T''(n_1),T''(n_2),\ldots$ of numbers such that
$n_0 < n_1 < n_2 < \ldots$ and 
$T''(n_0) < T''(n_1) < T''(n_2) < \ldots$,
which means that for almost all $n$,
$W_{P(T''[n])}$ contains the infinite set 
$\{T''(n_0),T''(n_1),T''(n_2),\ldots\}$.
Hence $P$ is a \WeakApprox\ learner.
Note that $P$ is also a \BCf\ learner
as it always outputs cofinite sets.

Now assume for a contradiction that for some 
$n$ and learner $Q$, $Q$ \BCn\ learns the class of
all cofinite sets.  Since $Q$ \BCn\ learns
$\natnum$, there is a $\sigma\in(\natnum\cup\{\#\})^*$
such that for all $\tau\in(\natnum\cup\{\#\})^*$,
$|\natnum - W_{Q(\sigma\tau)}| \leq n$.
Now choose some cofinite $L$ such that
$range(\sigma)\subset L$ and 
$|\natnum - L| \geq 2n+1$.
Since $Q$ must \BCn\ learn
$L$, there exists some $\theta\in(L\cup\{\#\})^*$
such that $|L \triangle W_{Q(\sigma\theta)}| \leq n$.
But $|\natnum - L| - |\natnum - W_{Q(\sigma\theta)}| \leq
|L \triangle W_{Q(\sigma\theta)}| \leq n$, and
so by the definition of $\sigma$, 
$|\natnum - L| \leq n + |\natnum - W_{Q(\sigma\theta)}|
\leq n+n = 2n$, contradicting the definition of $L$.
Therefore the class of all cofinite sets has no
\BCn\ learner for any $n$.          

Assume now that the set $L$ to be learnt is approximated with
parameter set $W$. Given an approximate learner $M$ for
this class, one can construct inductively a text $T$
such that either the text is for some set $L-\{w\}$ and
it conjectures almost always that $w$ is in the set to be
learnt or the text is for $L$ while there are infinitely
many conjectures which do not contain $W$ as a subset.

The idea is to construct the text $T$ step by step by
starting in (a) below and by alternating between (a)
and (b) as needed:

(a) Select a $w \in L \cap W$ not contained in the
part of the text constructed so far and add
to the part of the text the elements of $L-\{w\}$ in
ascending order until the learner $M$ on the so far
constructed initial segment conjectures a set not containing $w$;

(b) Append to the so far constructed part of the text
all elements of $L$ up to the element $w$ (inclusively)
and go back to step (a).

This gives then a text $T$ with the desired properties:
if the learner eventually stays in (a) forever, it is wrong on $w$
considered when it the last time goes into (a);
if the learner goes to (b) infinitely often, the text
$T$ is for $L$ while the learner $M$ conjectures infinitely
often sets which are not supersets of $W$. Thus there is
no approximate learner for the class of all cofinite sets.~\qed

\medskip
\noindent The following result shows that
weak approximate learning is quite powerful. 

\begin{thm}\label{infweakappro}
The class of all infinite sets is \ConsWeakApprox\ learnable. 
\end{thm}

\proof
Consider the learner $M$ which conjectures on
input $\sigma$ the set
$$
   W_{M(\sigma)} = range(\sigma) \cup \{x: \forall y \in range(\sigma)\,
     [x > y]\}
$$
and consider any text $T$ for an infinite set.
Let $S = \{x \in range(T)$: when $x$ appears first in $T$,
         no larger datum of $T$ has been seen so far$\}$.
Note that the set $S$ is infinite. Now all conjectures
$M(T[n])$ are a superset of $S$: if an $x \in S$ has
not yet appeared in $T[n]$ then all members of
$range(T[n])$ are smaller than $x$ and $x \in W_{M(T[n])}$
else $x$ has already appeared in $T[n]$ and
is therefore also in $range(T[n])$. Furthermore, if
$x \notin range(T)$ then almost all $n$ satisfy
$\max(range(T[n])) > x$ and therefore $x \notin W_{M(T[n])}$,
thus for every $x$ almost all hypotheses $W_{M(T[n])}$ are
correct at $x$.~\qed

\medskip
\noindent Unfortunately, the weakly approximate
learning property of any class of infinite
sets may be lost if finite sets are added
to the target class. 

\begin{prop}\label{goldclass}
Gold's class consisting of the set of natural numbers and all
sets $\{0,1,\ldots,m\}$ is not \WeakApprox\ learnable.
\end{prop}

\proof
Make a text $T$ where $T(0) = 0$ and iff the $n$-th hypothesis
of the learner contains $T(n)+1$ then $T(n+1) = T(n)$
else $T(n+1) = T(n)+1$.

In the case that the text $T$ is for a finite set with maximum $m$
then $T(n) = m$ for almost all $n$ and the $n$-th hypothesis
contains $m+1$ for almost all $n$; thus the approximations are
in the limit false at $m+1$.

In the case that the text $T$ is for the set of all natural numbers
then consider any $m>0$ and consider the first $n$ such that $T(n+1) =
m$. Then the $n$-th hypothesis does not contain $m$.  Therefore, one
can conclude that for every $m$ there is an $n \geq m$ such that the
$n$-th hypothesis is conjecturing $m$ not to be in the set to be
learnt although the set to be learnt is the set of all natural
numbers. In particular there is no infinite set on which from some
time on all approximations are correct.

Thus the class considered is not weakly approximately learnable.~\qed

\medskip
\noindent
It may be observed that in the proof of Theorem \ref{infweakappro},
the parameter sets $S$ with respect to which the learner
$M$ approximates the class of all infinite sets may not 
necessarily be r.e.\ (or be of any fixed Turing degree).  This
motivates the question of whether or not the class
of all infinite sets is still weakly approximately
learnable if one restricts the class of parameter
sets in Definition \ref{approxdef} to some countable family.   

\begin{defn}
For any sets $L$ and $W$, where $W$ is infinite,
and any text $T$ for $L$, say that a
recursive learner $M$ \emph{weakly approximately (\WeakApprox) 
learns $L$ via $W$ on $T$} 
iff for all finite variants $V$ of $W$, 
it holds that for almost all hypotheses $H$ of $M$
on $T$, $H \cap V = L \cap V$.
For any class $\mathcal{W}$ of infinite sets, 
a class $S$ of sets is \emph{weakly approximately (\WeakApprox)
learnable via $\mathcal{W}$} iff there
is a recursive learner $M$ such that for every
$L \in S$ and every text $T$ for $L$,
$M$ \WeakApprox\ learns $L$ 
via some $W \in \mathcal{W}$ on $T$.
\end{defn}

\begin{prop}
For any countable class $\mathcal{W}$ of
infinite sets, the class of all cofinite 
sets is not \WeakApprox\ learnable via $\mathcal{W}$. 
\end{prop}

\proof
Suppose $M$ is a recursive learner that 
weakly approximately learns all cofinite sets
via some countable class $\mathcal{W}$
of infinite sets.  First, note that there exist 
$\sigma\in\natnum^*$ and $V \in\mathcal{W}$
such that for all $\tau\in\natnum^*$,  
$V \subseteq W_{M(\sigma\tau)}$.  For, assuming
otherwise, one can build a text $T$ for $\natnum$
as follows.  Let $V_0,V_1,V_2,\ldots$ be a one-one
enumeration of $\mathcal{W}$,
and set $T_0 = \epsilon$, where $T_s$ denotes 
the text prefix built until stage $s$.
Let $m_s$ be the minimum number not
contained in $range(T_s)$, and find
strings $\eta_0,\eta_1,\ldots,\eta_s$
such that for all $i\in\{0,\ldots,s\}$,
$V_i \not\subseteq W_{M(T_s\circ m_s\eta_0
\ldots\eta_i)}$; by assumption, such
strings $\eta_0,\eta_1,\ldots,\eta_s$
must exist.  Let $T = \lim_s T_s$.
$T$ is a text for $\natnum$; furthermore,
for any $V_l \in \mathcal{W}$,
$V_l \not\subseteq W_{M(T[s+1])}$
for infinitely many $s$, so that
$M$ does not weakly approximately
learn $\natnum$ via $V_l$ on $T$.

Now fix $\sigma\in\natnum^*$ and
$V \in \mathcal{W}$ such
that for all $\tau\in\natnum^*$,
$V \subseteq W_{M(\sigma\tau)}$.
As $V$ is infinite, one can choose some 
$w \in V - range(\sigma)$.  Let
$T'$ be a text for $\natnum - \{w\}$
that extends $\sigma$.  Then 
$M$ conjectures a set containing
$w$ on almost all text prefixes
of $T'$, which shows that it
cannot weakly approximately
learn $\natnum - \{w\}$. 
In conclusion, the class of all
cofinite sets is not weakly
approximately learnable via
$\mathcal{W}$.~\qed

\begin{thm} \label{th:bcstarweakapprox}
If  $\mathcal{C}$ is \BCf\ learnable then $\mathcal{C}$ is
\WeakApprox\ learnable.
\end{thm}

\proof
By Theorem \ref{infweakappro}, there is a learner $M$ that weakly approximates the class
of all infinite sets. Let $O$ be a \BCf\ learner for
$\mathcal{C}$.  Now the new learner $N$ is given as follows:
On input $\sigma$, $N(\sigma)$ outputs an index of the following set
which first enumerates $range(\sigma)$ and then searches for
some $\tau$ that satisfies the following conditions:
(1) $range(\tau) = range(\sigma)$; (2)
$|\tau| = 2*|range(\sigma)|$;
(3) $W_{O(\tau \#^s)}$ enumerates at least $|\sigma|$ many elements
for all $s \leq |\sigma|$.
If all three conditions are met then the set contains also all
elements of $W_{M(\sigma)}$.
If $L \in C$ is finite then for every $\tau$ of length $2*|L|$ with
range $L$, the learner outputs on some input $\tau \#^{s_\tau}$ a
finite set with $c_\tau$ many elements. As there are only finitely
many such $\tau$, there is an upper bound $t$ of all $c_\tau$ and
$s_\tau$.  Then it follows from the construction that the learner $N$
on any input $\sigma$ with $range(\sigma) = L$ and $|\sigma| \geq t$
outputs a hypothesis for the set $L$, as the corresponding $\tau$
cannot be found. Thus $N$ weakly approximately learns~$L$.

If $L \in \mathcal{C}$ is infinite then there is a locking sequence
$\gamma \in L^*$ for $L$ such that $O(\gamma \eta)$ conjectures an
infinite set whenever $\eta \in L^*$. It follows for all $\sigma$ with
$range(\gamma) \subseteq range(\sigma)$
and $|range(\sigma)| > |\gamma|$ that $N(\sigma)$ considers also
a $\tau$ which is an extension of $\gamma$ in its algorithm and
which therefore meets all three conditions, thus $N(\sigma)$
will conjecture a set consisting of the union of
$range(\sigma)$ and $W_{M(\sigma)}$. As adding $range(\sigma)$
to the hypothesis $W_{M(\sigma)}$ cannot make $W_{N(\sigma)}$ to
be incorrect at any $x$ where $W_{M(\sigma)}$ is correct,
it follows that also $N$ is weak approximately learning $L$.
Thus, by case distinction, $N$ is a weak approximate learner for
$\mathcal{C}$.~\qed

\section{Combining Partial Language Learning With Variants of
Approximate Learning}

This section is concerned with the question whether partial learners
can always be modified to approximate the target language in the
models introduced above.

\subsection{Finitely Approximate Learning}

\noindent
The first results demonstrate the power of the model of finitely
approximate learning: there is a partial learner that finitely
approximates every r.e.\ language.

\begin{thm}\label{finapproxpart}
The class of all r.e.\ sets is \FinApproxPart\ learnable.
\end{thm}

\proof
Let $M_1$ be a partial learner of all
r.e.\ sets.  Define a learner $M_2$
as follows.  Given a text $T$,
let $e_n = M_1(T[n+1])$ for all $n$.  
On input $T[n+1]$, $M_2$ determines the 
finite set $D = range(T[n+1]) \cap \{0,\ldots,m\}$, 
where $m$ is the minimum $m \leq n$ with 
$e_m = e_n$.  $M_2$ then outputs a canonical index for
$D \cup (W_{e_n} \cap \{x: x > m\})$.

Suppose $T$ is a text for some r.e.\ set $L$.
Then there is a least $l$ such that $M_1$
on $T$ outputs $e_l$ infinitely often and
$W_{e_l} = L$.  Furthermore, there is a least 
$l'$ such that for all $l''> l'$, $D_L = range(T[l''+1])
\cap \{0,\ldots,l\} = L \cap \{0,\ldots,l\}$.
Hence $M_2$ will output a canonical index
for $L = D_L \cup (W_{e_l} \cap \{x: x > l\})$
infinitely often.  On the other hand, since,
for every $h$ with $e_h \neq e_l$ and
$e_h \neq e_{h'}$ for all $h' < h$, $M_1$
outputs $e_h$ only finitely often, $M_2$
will conjecture sets of the form
$D' \cup (W_{e_h} \cap \{x: x > h\})$
only finitely often.  Thus $M_2$ partially
learns $L$. 

To see that $M_2$ is also a finitely approximate
learner, consider any number $x$.  Suppose that 
$M_1$ on $T$ outputs exactly one index $e$ infinitely 
often; further, $W_e = L$ and $j$ is the least index
such that $e_j = e$.  Let $s$ be sufficiently large so 
that for all $s' > s$, $range(T[s'+1]) \cap \{0,\ldots,\max(\{x,j\})\} 
= L \cap \{0,\ldots,\max(\{x,j\})\}$.  First, assume that $M_1$ 
outputs only finitely many distinct indices on $T$.  
It follows that $M_1$ on $T$ converges to $e$.  
Thus $M_2$ almost always outputs a canonical 
index for $(L \cap \{0,\ldots,j\}) \cup (W_{e_j} \cap
\{y: y > j\})$, and so it approximately
learns $L$.  Second, assume that $M_1$ outputs
infinitely many distinct indices on $T$.
Let $d_1,\ldots,d_x$ be the first $x$
conjectures of $M_1$ that are pairwise
distinct and are not equal to $e$.  There
is a stage $t > s$ large enough so that 
$e_{t'} \notin \{d_1,\ldots,d_x\}$ for all
$t' > t$.  Consequently, whenever $t' > t$,
$M_2$ on $T[t'+1]$ will conjecture a set $W$
such that $W \cap \{0,\ldots,x\} = L \cap
\{0,\ldots,x\}$.  This establishes that $M_2$
finitely approximately learns any r.e.\ set.~\qed

\medskip
\noindent
It may be observed in the proof of Theorem \ref{finapproxpart}
that if $M_1$ is a confident partial
learner of some class $\mathcal{C}$, then $M_2$
confidently partially as well as finitely approximately
learns $\mathcal{C}$.  This observation leads to the next
theorem.

\begin{thm}\label{finapproxconfpart}
If $\mathcal{C}$ is \ConfPart\ learnable, then
$\mathcal{C}$ is \FinApproxConfPart\ learnable.
\end{thm}

\medskip
\noindent
Gao, Jain and Stephan \cite{consvpart} showed that consistently
partial learners exist for all and only the subclasses of uniformly
recursive families; the next theorem
shows that such learners can even be finitely 
approximate at the same time, in addition to being \emph{prudent}.

\begin{thm}\label{urecfinapprox}
If $\mathcal{C}$ is a uniformly recursive
family, then $\mathcal{C}$
is \FinApproxCons- \Part\ learnable by a prudent learner.
\end{thm}

\proof
Let $\mathcal{C} = \{L_0,L_1,L_2,\ldots,\}$ be a
uniformly recursive family.  On text $T$,
define $M$ at each stage $s$ as follows:
\begin{quote}
If there are $x \in {\mathbb N}$ and $i \in \{0,1,\ldots,s\}$ such that
\begin{itemize}
\item
$range(T[s+1])-range(T[s]) = \{x\}$,
\item
$range(T[s+1]) \subseteq L_i \cup \{\#\}$ and
\item
$range(T[s+1]) \cap \{0,\ldots,x\} = L_i \cap \{0,\ldots,x\}$
\end{itemize}
Then $M$ outputs the least such $i$ \\
Else $M$ outputs a canonical index for $range(T[s+1]) -\{\#\}$.
\end{quote}
The consistency of $M$ follows directly by construction.
If $T$ is a text for a finite set then the ``Else-Case''
will apply almost always and $M$ converges to
a canonical index for $range(T)$. Now consider that
$T$ is a text for some infinite set $L_m\in\mathcal{C}$
and $m$ is the least index of itself.
Let $t$ be large enough so that for all $t' > t$, 
all $x \in L - range(T[t+1]) - \{\#\}$ and all $j < m$,
$L_j \cap \{0,\ldots,x\} \neq
  range(T[t'+1]) \linebreak[3] \cap \linebreak[3] \{0,\ldots,x\}$.   
There are infinitely many stages $s > \max(\{t,m\})$ at which
$T(s) \notin range(T[s]) \cup \{\#\}$ and
$range(T[s+1]) \cap \{0,\ldots,T(s)\} = L \cap
\{0,\ldots,T(s)\}$.  At each of these stages,
$M$ will conjecture $L_m$.  Thus $M$ conjectures
$L_m$ infinitely often.  Furthermore, for
every $x$ there is some $s_x$ such that for
all $y \in L - range(T[s_x+1])$, it holds
that $y > x$.  Thus whenever $s' > s_x$,
$M$'s conjecture on $T[s'+1]$ agrees with $L$ 
on $\{0,\ldots,x\}$.  $M$ is therefore a 
finitely approximate learner, implying that
it never conjectures any incorrect index  
infinitely often.~\qed 

\medskip
\noindent
Proposition \ref{urecfinapprox} and \cite[Theorem 18]{gaostezill13} together give 
the following corollary.

\begin{cor}\label{finapproxconspart}
If $\mathcal{C}$ is \ConsPart\ learnable,
then $\mathcal{C}$ is \FinApproxConsPart\ learnable by a prudent learner.
\end{cor}

\noindent
The following result shows that also conservative partial learning
may always be combined with finitely approximate learning.

\begin{thm}\label{finapproxconsvpart}
If $\mathcal{C}$ is \ConsvPart\ learnable, then
$\mathcal{C}$ is \FinApproxConsvPart\ \linebreak[4]learnable. 
\end{thm}

\proof
Let $M_1$ be a \ConsvPart\ learner for $\mathcal{C}$,
and suppose that $M_1$ outputs the sequence of conjectures
$e_0,e_1,\ldots$ on some given text $T$.
The construction of a new learner $M_2$ is similar
to that in Theorem \ref{finapproxpart}; however,
one has to ensure that
$M_2$ does not output more than one index that is either
equal to or a proper superset of the target language. 
On input $T[s+1]$, define $M_2(T[s+1])$ as follows.
\begin{enumerate}
\item If $range(T[s+1]) \subseteq \{\#\} $ then output
a canonical index for $\emptyset$ else go to~2.
\item Let $m \leq s$ be the least number such that $e_m = e_s$.
If $W_{e_s,s} \cap \{0,\ldots,m\} = range(T[s+1]) \cap \{0,\ldots,m\}
= D$ then output a canonical index for $D \cup (W_{e_m} \cap
\{x:x > m\})$ else go to 3.  
\item If $s \geq 1$ then output $M_2(T[s])$ else 
output a canonical index for $\emptyset$. 
\end{enumerate}
Suppose that $T$ is a text for some $L\in\mathcal{C}$.
Without loss of generality, assume that $L \neq \emptyset$;
if $L = \emptyset$, then $M_2$ will always output a canonical
index for $\emptyset$.  $M_1$ on $T$ outputs exactly one 
index $e_h$ infinitely often, where $W_{e_h} = L$ and 
$e_{h'} \neq e_h$ for all $h' < h$.  Let $s$ be the least 
stage at which $range(T[s+1]) \cap \{0,\ldots,h\} = L \cap \{0,\ldots,h\}
= W_{e_h,s} \cap \{0,\ldots,h\}$.  Then for all $s' \geq s$
such that $e_{s'} = e_h$, step 2.\ will apply, so
that $M_2$ outputs a canonical index $g$ for 
$(L \cap \{0,\ldots,h\}) \cup (W_{e_h} \cap \{x:x > h\}) = L$.
Since there are infinitely many such $s'$, $M_2$
will output $g$ infinitely often.  Consider any
other set of the form $F \cup (W_{e_l} \cap \{x:x>l\})$
that $M_2$ may conjecture at some stage $t$, where $l \neq h$
and $e_{l'} \neq e_l$ for all $l' < l$.  By construction, $F$
is equal to $W_{e_l,t} \cap \{0,\ldots,l\}$.  Thus
$F \cup (W_{e_l} \cap \{x:x>l\}) \subseteq W_{e_l}$,
and so by the partial conservativeness of $M_1$, 
$L \not\subseteq F \cup (W_{e_l} \cap \{x:x>l\})$.
If $M_2$ conjectures some set of the form
$G \cup (W_{e_h} \cap \{x:x>h\})$, where
$G \neq L \cap \{0,\ldots,h\}$, then there is some
$y \in L - (G \cup (W_{e_h} \cap \{x:x>h\}))$,
and so $L \not\subseteq G \cup (W_{e_h} \cap \{x:x>h\})$.
Furthermore, $L \not\subseteq\emptyset$.  Therefore $M_2$
outputs exactly one index for a set that contains $L$,
and $M_2$ outputs this index infinitely often.
To show that $M_2$ outputs any incorrect index only
finitely often, it is enough to show that it finitely
approximately learns $L$.  

Consider any $x$.  If $M_1$ on $T$ outputs only finitely 
many distinct indices, then one can argue as in Theorem \ref{finapproxpart}
that $M_2$ converges on $T$ to $g$.  Suppose that
$M_1$ on $T$ outputs infinitely many
distinct indices.  Let $s$ be the least stage at
which $range(T[s+1])\cap\{0,\ldots,x\} = L \cap \{0,\ldots,x\}$.  
Let $d_1,\ldots,d_x$ be $x$ pairwise distinct indices 
of $M_1$ on $T$, none of which is equal to $e_h$.
Then there is a least stage $t > s$ such that
$M_2(T[t+1]) = g$ and for all $t' > t$, $e_{t'}\notin
\{d_1,\ldots,d_x\}$.  Thus on any $T[t'+1]$ with $t' > t$,
$M_2$ either outputs $g$ or conjectures a set $W$
such that $W \cap \{0,\ldots,x\} = L \cap \{0,\ldots,x\}$.
Therefore $M_2$ is both a finitely approximate and
a conservatively partial learner of $\mathcal{C}$.~\qed                    

\medskip
\noindent
Jain, Stephan and Ye \cite{jainsteye09} proved that
for uniformly r.e.\ classes, class-comprising 
explanatory learning is equivalent to \emph{uniform} 
explanatory learning; the latter means that one can
construct a numbering of partial-recursive
learners $M_0,M_1,\linebreak[3]M_2,\ldots$ such that
for any given r.e.\ numbering $H_0,H_1,H_2,\ldots$
of the target class $\mathcal{C}$ with $W_e = 
\{\langle d,x\rangle:x\in H_d\}$, the $e$-th 
learner explanatorily learns $\mathcal{C}$
with respect to $\{H_0,H_1,H_2,\ldots\}$.   
In particular, uniformly r.e.\ explanatorily
learnable classes are always explanatorily
learnable with respect to a class-preserving
hypothesis space.  The next theorem shows,
however, that none of the approximate learning 
criteria considered so far can be combined with 
class-preservingness.  Thus, in general, any 
successful approximation of languages must involve sets
not contained in the target hypothesis space.
An intuitive explanation for this is that a 
class-preserving learner may be incapable of 
recursively deciding, for any given finite set $D$,
whether there exists a language in the
target class that agrees with the current input
on $D$.   

\begin{thm}\label{approxnotclspresv}
There is a uniformly r.e.\ class that is 
\Approx\ learnable but not \ClsPresvFinApprox\ learnable.  
\end{thm}

\proof
Let $M_0,M_1,M_2,\ldots$ be an enumeration of
all partial-recursive learners.  For each $e$,
define a strictly increasing r.e.\ sequence
$x_{e,1},x_{e,2},\ldots$ as follows.  First,
for any given finite set $D$ and number $y \notin D$, 
let $\alpha_{D,y}$ denote the string
$1\circ 4\circ\ldots\circ 3y+1$, which
is a concatenation (in increasing order) of all numbers 
of the form $3z+1$ with $0 \leq z \leq y$ and 
$z \notin D$.  $x_{e,1}$ is defined to be the 
first number found (if such a number exists) such 
that for some $m_{e,1}$ with $x_{e,1} > m_{e,1}$, it 
holds that $\{3e,3x_{e,1}+1\} \subseteq W_{M_e(3e\circ\alpha_{\emptyset,m_{e,1}})}$.
Suppose that $x_{e,1},\ldots,x_{e,k}$ have
been defined.  $x_{e,k+1}$ is then defined
to be the first number found (if such a number
exists) such that for some $m_{e,k+1}$ with 
$x_{e,k+1} > m_{e,k+1} > x_{e,k}$, it holds that
$\{3e,3x_{e,k+1}+1\} \subseteq W_{M_e(3e\circ\alpha_{\{x_{e,i}:
1 \leq i\leq k\},m_{e,k+1}})}$. 

For each pair $\langle e,i\rangle$, define $L_{\langle e,i\rangle}$
according to the following case distinction. 

\medskip

\noindent \emph{Case (1): $x_{e,i}$ is defined for all $i$.}  
Set $L_{\langle e,0\rangle} = \{e\} \oplus
(\natnum - \{x_{e,i}: i\in\natnum\}) \oplus \emptyset$.
For each $j > 0$, set $L_{\langle e,j\rangle}
= \{e\} \oplus (\natnum - \{x_{e,i}: i < j \})
\oplus \{0\}$. 

\medskip

\noindent \emph{Case (2): There is a minimum $l$ such that
$x_{e,l}$ is undefined.}
Set $L_{\langle e,0\rangle} = \{e\} \oplus
(\{y:(l=1 \Rightarrow y < 0) \wedge 
(l > 1 \Rightarrow y < x_{e,l-1})\}-
\{x_{e,i}: i < l\}) \oplus \emptyset$.
For each $j$ with $1 \leq j \leq l-1$,
set $L_{\langle e,j\rangle} = \{e\} \oplus
(\natnum - \{x_{e,i}: i < j\}) \oplus \{0\}$.
Set $L_{\langle e,l\rangle} = \{e\} \oplus
(\natnum - \{x_{e,i}: i < l\}) \oplus \emptyset$.
For each $j \geq l+1$, set $L_{\langle e,j\rangle} = \emptyset$.

Set $\mathcal{C} = \{L_{\langle e,i\rangle}: e,i\in\natnum\}$.

Now it is shown that $\mathcal{C}$ is approximately
learnable with respect to a class-comprising
hypothesis space.  On input $\sigma$, the learner 
$M$ outputs a canonical index for
$\emptyset$ if $range(\sigma)$ does not contain any
multiple of $3$.  Otherwise, let $e$ be the minimum
number such that $3e\in range(\sigma)$; $M$ then checks 
whether or not $2 \in range(\sigma)$.
If $2 \in range(\sigma)$, $M$ searches
(with computational time bounded by $|\sigma|$) for 
the least $l$ (if such an $l$ exists) such that 
$3x_{e,l}+1\in range(\sigma)$; it then conjectures 
$L_{\langle e,l\rangle}$.  If no such $l$ exists,
$M$ conjectures $L_{\langle e,1\rangle}$.  
If $2 \notin range(\sigma)$,
$M$ searches for the minimum $l'$
such that $x_{e,l'}$ has not yet been defined 
at stage $|\sigma|$.  If $3x_{e,l'}+1 \notin 
range(\sigma)$, then $M$ conjectures $L_{\langle e,0\rangle}$.
If $3x_{e,l'}+1 \in range(\sigma)$, then
$M$ outputs an index $d$ such that
$$
W_d = \left\{ \begin{array}{ll}
range(\sigma) \cup \{3z+1: (l'=1 \Rightarrow 0 \leq z \leq s)
& \mbox{if $s > x_{e,l'}$ is the first step at} \\
\wedge (l' > 1 \Rightarrow x_{e,l'-1}+1 \leq z \leq s)\}
\cup L_{\langle e,0\rangle} & \mbox{which $x_{e,l'}$ is defined;} \\
range(\sigma) \cup \{3z+1: (l'=1 \Rightarrow z \geq 0) &
\mbox{if $x_{e,l'}$ is undefined.} \\
\wedge (l' > 1 \Rightarrow z \geq x_{e,l'-1}+1)\}\end{array}\right. 
$$
For the verification that $M$ approximately learns $\mathcal{C}$, 
suppose that $M$ outputs the sequence of conjectures 
$e_0,e_1,e_2,\ldots$ on text $T$.  Assume first that
$x_{e,i}$ is defined for all $i$.  If $T$ is a 
text for $L_{\langle e,0\rangle}$, then for almost all $n$, 
$W_{e_n}$ is a finite variant of $L_{\langle e,0\rangle}$; furthermore,
if $e_{j_0},e_{j_1},\ldots$ is the subsequence
of conjectures for which $W_{e_{j_i}} \neq L_{\langle e,0\rangle}$, 
then the sequence $y_0,y_1,y_2,\ldots$ of minimum
numbers such that $W_{e_{j_i}}(y_i) \neq
L_{\langle e,0\rangle}(y_i)$ is almost
always monotone increasing and contains
a strictly increasing subsequence.  In addition,
for almost all $i$, $W_{e_i}(y) = L_{\langle e,0\rangle}(y)$ 
for all $y$ contained in $L_{\langle e,0\rangle}$, 
which is an infinite set.  Hence $M$
approximately learns $L_{\langle e,0\rangle}$.
If $T$ is a text for $L_{\langle e,j\rangle}$
for some $j > 0$, then $2\in range(T)$ and
so $M$ will eventually identify $j$ as the minimum $l$ 
such that $3x_{e,l}+1 \in range(T)$.
Thus $M$ will converge to an index for
$L_{\langle e,j\rangle}$.  Next, assume that 
there is a minimum $l$ such
that $x_{e,l}$ is undefined.  If
$T$ is a text for $L_{\langle e,0\rangle}$,
then $M$ will in the limit identify
$l$ as the minimum $l'$ such that
$x_{e,l'}$ is undefined; thus, as
$3x_{e,l'}+1\notin range(T)$, $M$ on $T$
will converge to an index for $L_{\langle e,0\rangle}$.
If $T$ is a text for some nonempty $L_{\langle e,j\rangle}$
with $j > 1$, $M$ on $T$ will again
converge to an index for $L_{\langle e,j\rangle}$:
if $2 \in L_{\langle e,j\rangle}$, then
$M$ will eventually identify $j$ as the minimum
number $l$ such that $3x_{e,l}+1 \in range(T)$
and converge to indices for $L_{\langle e,j\rangle}$;
if $2 \notin L_{\langle e,j\rangle}$, 
then $3x_{e,j}+1 \in range(T)$
and the fact that $j$ is the minimum
number for which $x_{e,j}$ is undefined
together imply that $M$ on $T$ will converge
to indices for $L_{\langle e,j\rangle}$. 
By construction, $M$ converges to a canonical
index for $\emptyset$ on any text with
an empty range.  This completes the verification that
$M$ approximately learns $\mathcal{C}$.

It remains to show that $\mathcal{C}$ is not
\FinApprox\ learnable using a class-preserving
hypothesis space.  Assume that $M_e$ 
\ClsPresvFinApprox\ learns $\mathcal{C}$.
If there is a minimum $l$ such that
$x_{e,l}$ is undefined, then there
is a text $U$ for $L_{\langle e,l\rangle}$
on which $M_e$ almost always outputs
a conjecture that is different from
$L_{\langle e,l\rangle}$.  Since $M_e$
finitely approximates $L_{\langle e,l\rangle}$,
almost all of $M_e$'s hypotheses on
$U$ must contain $3e$.  But for all
$j > 0$ such that $j \neq l$, either 
$L_{\langle e,j\rangle} = \emptyset$
or $2 \in L_{\langle e,j\rangle}$.
As $2 \notin L_{\langle e,l\rangle}$
and $L_{\langle e,l\rangle}$ is infinite, 
while $L_{\langle e,0\rangle}$ is finite,
it follows that $M_e$, being a finitely
approximate learner, must almost always
conjecture a set different from any
$L_{\langle e,j\rangle}$ with $j \neq l$.
Hence $M_e$ is not a finitely approximate
learner of $L_{\langle e,l\rangle}$.
Suppose, on the other hand, that $x_{e,i}$
is defined for all $i$.  Then one can
build a text $U'$ for $L_{\langle e,0\rangle}$
on which $M_e$ infinitely often conjectures a set 
containing $2$; but since $2 \notin L_{\langle e,0\rangle}$,
it follows that $M_e$ does not finitely 
approximately learn $L_{\langle e,0\rangle}$.
This establishes that $\mathcal{C}$
is not \ClsPresvFinApprox\ learnable.~\qed

\medskip
\noindent
The main content of the following proposition may
be summed up as follows: the quality of the 
hypotheses issued by a \BCf\ learner may be
improved so that for any given finite set $D$,
the learner's hypotheses will eventually
agree with the target language on $D$.    

\begin{prop}\label{finapproxbcf}
If $\mathcal{C}$ is \BCf\ learnable, then $\mathcal{C}$
is \FinApproxBCf\ learnable.
\end{prop}

\proof
Given a \BCf\ learner $M$ of $\mathcal{C}$, one can 
make a new learner $N$ as follows.  On input $\sigma$,
$N$ conjectures $range(\sigma) \cup 
(W_{M(\sigma)} \cap \{z: z > |\sigma|\})$.
Suppose that $N$ is fed with a text $T$ for some
$L\in\mathcal{C}$.  $N$ is a \BCf\ learner
because it always conjectures finite
variants of $M$'s conjectures.  Furthermore, 
for every finite set $D$ there is some 
$s_D$ such that $s_D > \max(D)$ and $range(T[s]) \cap D
= L \cap D$ for all $s > s_D$.  It follows by construction
that for all $s > s_D$, $W_{N(T[s])} \cap D
= range(T[s]) \cap D = L \cap D$, and so
$N$ finitely approximately \BCf\ learns $L$.~\qed

\medskip
\noindent
The next two results consider combinations
of finite approximation and some learning models 
that permit finitely many anomalies.  It
is readily seen that the additional constraint
of finite approximation implies that any anomaly 
in the learner's hypotheses will eventually 
be corrected.

\begin{prop}
If $\mathcal{C}$ is \VacfFinApprox\ learnable,
then $\mathcal{C}$ is \Vac\ learnable.
\end{prop}

\begin{prop}
If $\mathcal{C}$ is \ExfFinApprox\ learnable,
then $\mathcal{C}$ is \Ex\ learnable. 
\end{prop}

\medskip
\noindent
As a side remark, \ConsvPartBC\ learning is only
as powerful as \ConsvEx\ learning; the following 
proposition establishes this fact.

\begin{prop}\label{consvpartbc}
If $\mathcal{C}$ is \ConsvPartBC\ learnable,
then $\mathcal{C}$ is \PrudConsvEx\ learnable.
\end{prop}

\proof
Note that on any text for some $L \in \mathcal{C}$, a 
\ConsvPartBC\ learner $M$ outputs exactly one index
$e$ with $W_e = L$; since $M$ is also a \BC\ learner,
this means that $M$ on $T$ converges to $e$ and
it never outputs a proper superset of $L$.  
By \cite[Theorem 29]{gaostezill13} and
\cite[Theorem 10]{consvpart}, $\mathcal{C}$
is \PrudConsvEx\ learnable.~\qed

\subsection{Weakly Approximate, Approximate and \BCf\ Learning}

\noindent
The next proposition
shows that Theorem~\ref{urecfinapprox} cannot
be improved and gives a negative answer
to the question whether partial
or consistent partial learning can be combined
with weakly approximate learning.

\begin{prop}\label{weakapproxpart}
The uniformly recursive class $\{A: A = {\mathbb N}$ or $A$ contains
all even and finitely many odd numbers or $A$ contains
finitely many even and all odd numbers$\}$ is
(a) \ConsWeakApprox\ learnable and (b) \ConsPart\
learnable, but not \WeakApproxPart\ learnable.
\end{prop}

\proof
That (a) can be satisfied follows from Theorem
\ref{infweakappro}; that (b) can be satisfied follows
from \cite[Theorem 18]{gaostezill13}.
Furthermore, one can easily make a text $T$ which makes sure that 
a given partial learner $M$ for the class does not also weakly 
approximate it. The idea is to define the text $T$ inductively 
as follows by going through the following loop: 
\begin{enumerate}
\item Let $n=0$;
\item As long as $M(T[n])$ does not conjecture a set which
      contains all even numbers and only finitely many odd
      numbers let $T(n)$ be the least even number not yet
      in the text and update $n=n+1$;
\item As long as $M(T[n])$ does not conjecture a set which
      contains all odd numbers and only finitely many even
      numbers let $T(n)$ be the least odd number not yet
      in the text and update $n=n+1$;
\item Go to Step 2.
\end{enumerate}
It is easy to see that as the learner is partial it cannot
get stuck in Step 2 or Step 3 forever, as it would not output
an index for $range(T)$ infinitely often in that case.
Hence it alternates between Steps 2 and 3 infinitely often
and will therefore alternating between sets containing all
even and only finitely many odd numbers and all odd and only
finitely many even numbers. Hence there is no infinite set
which is contained in almost all hypotheses; however, the range
of $T$ is the set of natural numbers and thus the learner is
not weakly approximating it.~\qed

\noindent
The next theorem shows that neither partial learning nor
consistent partial learning can be combined
with approximate learning.  In fact, it establishes
a stronger result: consistent partial learnability and 
approximate learnability are insufficient
to guarantee both partial and weakly approximate
learnability simultaneously.   

\begin{thm} \label{th:separation}
There is a class of r.e.\ sets with the following properties:

\noindent
(i) The class is not \BCf\ learnable;

\noindent
(ii) The class is not \WeakApproxPart\ learnable; 

\noindent
(iii) The class is \Approx\ learnable;

\noindent
(iv) The class is \ExKjump\ learnable.

\noindent
(iv) The class is \ConsPart\ learnable.

\end{thm}

\proof
The key idea is to diagonalise against a list $M_0,M_1,\ldots$ of learners
which are all total and which
contains for every learner to be considered a delayed version. This
permits to ignore the case that some learner is undefined on some input.


The class witnessing the claim consists of all sets $L_d$ such that
for each $d$,
either $L_d$ is $\{d,d+1,\ldots\}$ or $L_d$ is a subset built
by  the following diagonalisation procedure:
One assigns to each number $x \geq d$ a level $\ell(x)$. 
\begin{itemize}
\item If some set $L_{d,e} = \{x \geq d: \ell(x) \leq e\}$ is infinite
then let $L_d = L_{d,e}$ for the least such $e$
      and $M_d$ does not partially learn $L_d$
\item else let $L_d = \{d,d+1,\ldots\}$ and $M_d$ does not
      weakly approximate $L_d$.
\end{itemize}
The construction of the sets is inductive over stages. 
For
each stage $s=0,1,2,\ldots$:
\begin{itemize}
\item Let $\tau_e$ be a sequence of all $x \in \{d,d+1,\ldots,d+s-1\}$
      with $\ell(x) = e$ in ascending order;
\item If there is an $e < s$ such that $e$ has not been cancelled in
      any previous step and for each $\eta \preceq \tau_e$ the intersection
      $
         W_{M_d(\tau_0\tau_1\ldots\tau_{e-1}\eta),s} \cap
         \{y: d \leq y < d+s \wedge \ell(y) > e\}
      $
      contains at least $|\tau_e|$ elements
\begin{itemize}
\item Then choose the least such $e$ and
      let $\ell(d+s) = e$ and cancel
      all $e'$ with $e < e' \leq s$
\item Else let $\ell(d+s) = s$.
\end{itemize}
\end{itemize}
A text $T = \lim_e \sigma_e$ is defined
as follows (where $\sigma_0$ is the empty sequence):
\begin{itemize}
\item Let $\tau_e$ be the sequence of all $x$ with $\ell(x) = e$
      in ascending order;
\item If $\sigma_e$ is finite then let $\sigma_{e+1} = \sigma_e \tau_e$
      else let $\sigma_{e+1} = \sigma_e$.
\end{itemize}
In case some $\sigma_e$ are infinite, let $e$ be smallest such that
$\sigma_e$ is infinite. Then $T = \sigma_e$ and $L_d = L_{d,e}$ and
$T$ is a text for $L_d$. As $L_{d,e}$ is infinite, one can conclude
that
$$
   \forall \eta \preceq \sigma_e\, \forall c \,
         [|W_{M_d(\tau_0\tau_1\ldots\tau_{e-1}\eta)} \cap \{y: \ell(y) > e\}|
          \geq c]
$$
and thus $M_d$ outputs on $T$ almost
always a set containing infinitely many elements outside $L_d$;
so $M_d$ does neither partially learn $L_d$ nor \BCf\ learn $L_d$.

In case all $\sigma_e$ are finite and therefore
all $L_{d,e}$ are finite there must be
infinitely many $e$ that never get cancelled. Each such $e$ satisfies
$$
   \exists \eta \preceq \tau_e\,
         [W_{M_d(\tau_0\tau_1\ldots\tau_{e-1}\eta)} \cap \{y: \ell(y) > e\}
          \mbox{ is finite}]
$$
and therefore $e$ also satisfies
$
   \exists \eta \preceq \tau_e\,
         [W_{M_d(\tau_0\tau_1\ldots\tau_{e-1}\eta)} \mbox{ is finite}].
$ 
Thus $M_d$ outputs on the text $T$ for the cofinite set
$L_d = \{d,d+1,\ldots\}$ infinitely often a finite set and $M_d$ is
neither weakly approximately learning $L_d$
(as there is no infinite set on which almost all conjectures are correct)
nor $BC^*$-learning $L_d$. Thus claims (i) and (ii) are true.

Next it is shown that the class of all $L_d$ is approximately learnable
by some learner $N$. This learner $N$ will on a text for $L_d$ eventually
find the minimum $d$ needed to compute the function $\ell$.
Once $N$ has found this $d$, $N$ will on each input $\sigma$
conjecture the set
$$
   W_{N(\sigma)} = \{x: x \geq \max(range(\sigma)) \vee
                     \exists y \in range(\sigma)\,[\ell(x) \leq \ell(y)]\}
$$
In case $L_d = L_{d,e}$ for some $e$,  $L_{d,e}$
is infinite, and for each text for $L_{e,d}$, almost all prefixes $\sigma$
of this text satisfy $\max\{\ell(y): y \in range(\sigma)\} = e$ and
$L_{d,e} \subseteq W_{N(\sigma)}$. So almost all conjectures are
correct on the infinite set $L_d$ itself.
Furthermore, $W_{N(\sigma)}$ does not contain any $x <
\max(range(\sigma))$ with $\ell(x) > e$, hence $N$ eventually becomes
correct also on any $x \notin L_{d,e}$ and therefore $N$ approximates
$L_{d,e} = L_d$.

In case $L_d = \{d,d+1,\ldots\}$, all
$L_{d,e}$ are finite. Then consider the infinite set
$S = \{x: \forall y > x\,[\ell(y) > \ell(x)]\}$. Let $x \in S$ and
consider any $\sigma$ with
$\min(range(\sigma))=d$.  If $x \geq \max(range(\sigma))$ then $x \in
W_{N(\sigma)}$.  If $x < \max(range(\sigma))$ then
$\ell(\max(range(\sigma))) \geq \ell(x)$
and again $x \in W_{N(\sigma)}$. Thus $W_{N(\sigma)}$ contains $S$. 
Furthermore, for all $x \geq d$ and sufficiently long prefixes
$\sigma$ of the text, $\ell(\max(range(\sigma))) \geq \ell(x)$ and
therefore all $x \in W_{N(\sigma)}$ for almost all prefixes $\sigma$
of the text. So again $N$ approximates $L_d$.
Thus claim (iii) is true.

Furthermore, there is a $K'$-recursive learner $O$ which explanatorily
learns the class. On input $\sigma$ with at least one element in
$range(\sigma)$, the learner determines $d = \min(range(\sigma))$.
If there is now some $e \leq |\sigma|$ such that $L_{d,e}$ is
infinite then $O$ conjectures $L_{d,e}$ for the least such $e$
else $O$ conjectures $\{d,d+1,\ldots\}$. It is easy to see
that these hypotheses converge to the set $L_d$ to be learnt:
eventually the minimum of the range of each input is $d$.
In the case that $L_d = L_{d,e}$ for some $e$ this $e$ is
detected whenever the input is longer than $e$ and therefore
the learner converges to $L_{d,e}$. In the case that all $L_{d,e}$
are finite, the learner almost always outputs the same hypothesis
for $\{d,d+1,\ldots\}$. Thus $O$ is a \ExKjump\ learner
and condition (iv) is true.

It remains to show that the class is \ConsPart\ learnable.
This follows from the fact that the class is a subclass
of the uniformly recursive family $\mathcal{U}=\{L_{e,d}\}_{e,d\in\natnum}
\cup\{\{d+x:x\in\natnum\}:d\in\natnum\}$.  To see that
$\mathcal{U}$ is uniformly recursive, it may be
observed from the construction of $L_{e,d}$ that
for each $d$, $\ell(x)$ is defined for all $x \geq d$;
each of these values, moreover, can be calculated 
effectively.  Thus one can uniformly decide for all
$d,e$ and $y$ whether or not $y \geq d$ and $\ell(y)\leq e$,
that is, whether or not $y \in L_{e,d}$.  Consequently,
by \cite[Theorem 18]{gaostezill13}, the given class is consistently
partially learnable, as required.~\qed

\medskip
\noindent
The next result separates \ConsApproxPart\ learning from \BCf\ learning. 

\begin{prop}\label{propsep:consapproxpartnotbcf}
The class $\mathcal{C} = \{\natnum\} \cup \{\{0,\ldots,e\} \cup \{2x: 
2x > e\}: e \in \natnum\}$ is \ConsApproxPart\ learnable
but not \BCf\ learnable.
\end{prop}

\proof
Make a learner $M$ as follows.  On input $\sigma$, if $range(\sigma)
-range(\sigma') = \{x\}$ for some odd number $x$, then $M$ outputs
a canonical index for $\natnum$.  Otherwise, $M$ determines the
maximum odd number $d$ (if such a $d$ exists) such that 
$d \in range(\sigma)$, and outputs a canonical
index for $\{y: y \leq d\} \cup \{2z: 2z > d\}$.  If no such $d$ 
exists, then $M$ outputs a canonical index for the set of all
even numbers.  Note that $M$ is consistent by construction.  
If $M$ is fed with a text $T$ for some set $L = \{0,\ldots,e\} 
\cup \{2x: 2x > e\}$, then there is a least $s$ such that
$\{0,\ldots,e\} \subseteq range(T[s])$.  
Thus for all $s' \geq s$, $M$ will output a canonical index 
for $L$ and so it explanatorily learns $L$.  If $M$ is fed with a text
for $\natnum$, then it will output a canonical index
for $\natnum$ at all stages where a new odd number appears; 
that is, it will output a canonical index for $\natnum$ infinitely often.
Furthermore, since $M$'s conjecture at every stage contains
the set of all even numbers, and $\{0,\ldots,f\}$
is contained in almost all of $M$'s conjectures for
every $f$, $M$ is an approximate learner, which implies
that it never outputs any incorrect index infinitely
often.  Hence $M$ \ConsApproxPart\ learns $\mathcal{C}$.

To see that $\mathcal{C}$ is not \BCf\ learnable, note that
if some learner $N$ \BCf\ learns $\natnum$, then there
is a $\sigma\in(\natnum\cup\{\#\})^*$ such that
for all $\tau\in(\natnum\cup\{\#\})^*$, $W_{N(\sigma\tau)}$
is cofinite: otherwise, one can build a text $T'$ for $\natnum$
such that $N$ on $T'$ outputs a coinfinite set infinitely
often, contradicting the fact that $N$ \BCf\ learns $\natnum$.
If $range(\sigma) = \emptyset$, let $d = 0$; otherwise,
let $d = \max(range(\sigma))$. 
Then one can extend $\sigma$ to a
text $\sigma\circ T''$ for $L' = \{0,\ldots,d\} \cup \{2z: 2z > d\}$.
By the choice of $\sigma$, $N$ on $\sigma\circ T''$ almost always
outputs a cofinite set, and so it does not even partially
learn $L'$.  Therefore $\mathcal{C}$ is not \BCf\ learnable.~\qed

\begin{rem}\label{prop:sepapproxbcfpartcons}
Note that \ApproxBCfPart\ learning cannot in general be 
combined with consistency; for example, consider the
class $\{K\}$, which is finitely learnable but cannot
be consistently learnt because $K$ is not recursive 
\cite[Theorem 18]{gaostezill13}.
\end{rem}

\medskip
\noindent
While the preceding negative results suggest that approximate and weakly
approximate learning imposes constraints that are too stringent for
combining with partial learning, at least partly positive results can
be obtained. For example, the following theorem shows that \ConsvPart\
learnable classes are \ApproxPart\ learnable (thus dropping only the
conservativeness constraint) by \BCf\ learners. This considerably improves an
earlier result by Gao, Stephan and Zilles \cite{gaostezill13} which states
that every \ConsvPart\ learnable class is also \BCf\ learnable.

\begin{thm}\label{consvpartimpliesbcfapprox}
If $\mathcal{C}$ is \ConsvPart\ learnable then
$\mathcal{C}$ is \ApproxPart\ learnable by a \BCf\ learner.
\end{thm}

\proof
Let $M$ be a \ConsvPart\ learner for $\mathcal{C}$. For a text $T$ for
a language $L \in
\mathcal{C}$, one considers the sequence $e_0,e_1,\ldots$ of distinct
hypotheses issued by $M$; it contains one correct hypothesis while all
others are not indices of supersets of $L$. For each hypothesis $e_n$ one
has two numbers tracking its quality: $b_{n,t}$ is the maximal $s
\leq n+t$ such that all $T(u)$ with $u<s$ are in $W_{e_n,n+t} \cup
\{\#\}$ and $a_{n,t} = 1+\max\{b_{m,t}: m < n\}$.

Now one defines the hypothesis set $H_{e_n,\sigma}$ for any sequence
$\sigma$. Let $e_{n,0},e_{n,1},\ldots$ be a
sequence with $e_{n,0} = e_n$ and $e_{n,u}$ be the $e_m$ for the
minimum $m$ such that $m=n$ or $W_{e_m}$
has enumerated all members of $range(\sigma)$ within $u+t$ time steps.
 The set $H_{e_n,\sigma}$ contains all $x$ for which there is a $u
\geq x$ with $x \in W_{e_{n,u}}$.

An intermediate learner $O$ now conjectures some canonical index
of a set $H_{e_n,\sigma}$ at least $k$ times iff there is a $t$
with $\sigma = T(0)T(1)\ldots T(a_{n,t})$ and $b_{n,t} > k$.
Thus $O$ conjectures $H_{e_n,\sigma}$ infinitely often
iff $W_{e_n}$ contains $range(T)$ and $a_{n,t} = |\sigma|$
for almost all $t$.

If $e_n$ is the correct index for the set to be learnt then, by
conservativeness, the sets $W_{e_m}$ with $m < n$ are not supersets of
the target set.  So the values
$b_{m,t}$ converge which implies that $a_{n,t}$ converges to
some $s$. It follows that for the prefix $\sigma$
of $T$ of length $s$, the canonical index of $H_{e_n,\sigma}$
is conjectured infinitely often while no other index is conjectured
infinitely often. Thus $O$ is a partial learner. Furthermore, for all
sets $H_{e_m,\tau}$ conjectured after $a_{n,t}$ has reached its final
value $s$, it holds that the $e_{m,u}$ in the construction of
$H_{e_m,\tau}$ converge to $e_n$. Thus $H_{e_m,\tau}$
is the union of $W_{e_n}$ and a finite set. Hence $O$ is a
\BCf\ learner. To guarantee the third condition on approximate
learning,  $O$ will be translated into
another learner $N$.

Let $d_0,d_1,\ldots$ be the sequence of $O$ output on the text $T$. 
Now $N$ will copy this sequence but with some delay. Assume that
$N(\sigma_k) = d_k$ and $\sigma_k$ is a prefix of $T$. Then $N$ will
keep the hypothesis $d_k$ until the current prefix $\sigma_{k+1}$
considered satisfies either 
$range(\sigma_{k+1}) \not\subseteq range(\sigma_k)$ or
$W_{d_k,|\sigma_{k+1}|} \neq range(\sigma_{k+1})$.

If $range(T)$ is infinite, the sequence of
hypotheses of $N$ will be the same as that
of $O$, only with some additional delay. Furthermore, almost all
$W_{d_n}$ contain $range(T)$, thus
the resulting learner $N$ learns $range(T)$ and is almost always
correct on the infinite set $range(T)$; in addition, $N$ learns $range(T)$
partially and is also \BCf.  If $range(T)$ is
finite, there will be some correct index that equals infinitely
many $d_n$. There is a step $t$ by which all elements of $range(T)$
have been seen in the text and enumerated into $W_{d_n}$.
Therefore, when the learner conjectures this correct index again, it
will never withdraw it; furthermore, it will replace eventually every
incorrect conjecture due to the comparison of the two sets. Thus the
learner converges explanatorily to $range(T)$ and is also in this case
learning $range(T)$ in a \BCf\ way, partially and approximately.
From the proof of Theorem \ref{finapproxpart}, one can see that $N$ may
be translated into a learner satisfying all the requirements
of \ApproxPart\ and \BCf learning.~\qed

\begin{exmp}\label{sepapproxconsv}
The class $\{\{e+d:d\in\natnum\}:e\in\natnum\} \cup 
\{\{e+d: e\in K-K_d\}:e\in\natnum\}$ is \Ex\ learnable
and hence \ApproxBCfPart\ learnable, but it is not
\ConsvPart\ learnable \cite[Theorem 29]{gaomthesis}.
\end{exmp}


\medskip
\noindent
Case and Smith \cite{Cas83} published Harrington's
observation that the class of recursive functions is 
\BCf\ learnable.  This result does not carry over
to the class of r.e.\ sets; for example, Gold's class 
consisting of the set of natural numbers and all
finite sets is not \BCf\ learnable.  In light of
Theorem \ref{recbcpartappro}, which established that the 
class of recursive functions can be \BCf\ and \Part\ 
learnt simultaneously, it is interesting to know 
whether \emph{any} \BCf\ learnable class of r.e.\ sets
can be both \BCf\ and \Part\ learnt at the same time.    
While this question in its general form remains open, 
the next result shows that \BCn\ learning is indeed combinable
with partial learning.

\begin{thm}\label{thm:bcnpart}
Let $n\in\mathbb{N}$. If $\mathcal{C}$ is \BCn\
learnable, then $\mathcal{C}$ is  \Part\ learnable by a \BCn\ learner.
\end{thm}

\proof
Fix any $n$ such that $\mathcal{C}$ is
\BCn\ learnable.  Given a recursive
\BCn\ learner $M$ of $\mathcal{C}$,
one can construct a new learner $N_1$
as follows.  First, let $F_0,F_1,F_2,\ldots$
be a one-one enumeration of all finite
sets such that $|F_i| \leq n$ for all
$i$.  Fix a text $T$, and let $e_0,e_1,e_2,\ldots$ 
be the sequence of $M$'s conjectures on $T$.

For each set of the form $W_{e_i} \cup F_j$
(respectively $W_{e_i} - F_j$), $N_1$ outputs
a canonical index for $W_{e_i} \cup F_j$
(respectively $W_{e_i} - F_j$) at
least $m$ times iff the following two conditions 
hold.
\begin{enumerate}
\item There is a stage $s > j$ for which
the number of distinct $x < j$ such that
either $x \in W_{e_i,s} \wedge
x \notin range(T[s+1])$ or
$x \in range(T[s+1]) \wedge x \notin 
W_{e_i,s}$ holds does not exceed $n$. 
\item There is a stage $t > m$ such that
for all $x < m$, $x \in W_{e_i,t} \cup F_j$
iff $x \in range(T[t+1])$ (respectively
$x \in W_{e_i,t} - F_j$ iff $x \in range(T[t+1])$).
\end{enumerate}
At any stage $T[s+1]$ where no set of the form
$W_{e_i} \cup F_j$ or $W_{e_i} - F_j$
satisfies the conditions above, or
each such set has already been output the
required number of times (up to the
present stage), $N_1$ outputs
$M(T[s+1])$.  

Suppose $T$ is a text
for some $L\in\mathcal{C}$.  Since
$M$ is a \BCn\ learner of $\mathcal{C}$,
it holds that for almost all $i$, there are at 
most $n$ $x$'s such that $W_{e_i}(x) \neq L(x)$.    
Furthermore, for all $j$ such that
$W_{e_j}(x) \neq L(x)$ for at least
$n+1$ distinct $x$'s, there is an
$l$ such that for all $l' > l$,
neither $W_{e_j} \cup F_{l'}$
nor $W_{e_j} - F_{l'}$ will satisfy
Condition 1.; thus, for any set
$S$ such that $S(x) \neq L(x)$ for
more than $n$ distinct values
of $x$, $N_1$ will conjecture $S$
only finitely often.  On the other hand, 
if there are at most $n$ distinct $x$'s such 
that $W_{e_i}(x) \neq L(x)$, then there 
is some $l$ such that either 
$L = W_{e_i} \cup F_l$ or $L = W_{e_i} - F_l$; 
consequently, either $W_{e_i} \cup F_l$ or $W_{e_i} - F_l$
will satisfy Conditions 1.\ and 2.\ for 
infinitely many $m$.  Hence $N_1$
is a \BCn\ learner of $L$ and it outputs 
at least one correct index for $L$ infinitely
often on any text for $L$.  Using
a padding technique, one can define a further 
learner $N$ that \BCnPart\ learns $\mathcal{C}$.~\qed

\medskip
\noindent
Theorems \ref{bcfwpart} and \ref{bcfinfinite} show that partial \BCf\
learning is possible for classes that can be \BCf\ learned by learners
that satisfy some additional constraints.  

\begin{thm}\label{bcfwpart}
Assume that $\mathcal{C}$ is \BCf\ learnable by a learner that
outputs on each text for any $L \in \mathcal{C}$ at least once a fully
correct hypothesis.
  Then $\mathcal{C}$ is \Part\ learnable by a \BCf\ learner.
\end{thm}

\proof
Let $M$ be given and on a text $T$, let $e_0,e_1,\ldots$
be the sequence of hypotheses by $M$. Now one can make
a learner $O$ which on input $T(0)T(1)\ldots T(n)$,
first computes $e_0,e_1,\ldots,e_n$
and then computes for every $e_m$ the quality $q_{m,n}$
which is the maximal number $y \leq n$ such that
for all $x \leq y$ the number $x$ has been enumerated
into $W_{e,n}$ iff $x \in \{T(0),T(1),\ldots,T(n)\}$.
In each step the learner $O$ outputs either the hypothesis
for the least $m$ such that either (a) $e_m$ has been output so far
less than $q_{m,n}$ times or (b) all $k \leq n$ satisfy that $e_k$ has
been output $q_{k,n}$ times and $q_{k,n} \leq q_{m,n}$.  One can see
that false hypotheses $e_m$ get output only
finitely often output while at least one correct hypotheses gets
output infinitely often; as all but finitely many hypotheses of $M$
are finite variants of $L$, the same is true for the modified learner
$O$.  By applying a padding technique, $O$ can be
converted to a learner $N$ which is at the same time
a \BCf\ learner and a partial learner.~\qed

\medskip
\noindent
The next definition gives an alternative
way of tightening the constraint
of \BCf\ learning. 

\begin{defn}
Let $\mathcal{C}$ be a class of r.e.\ sets.
A recursive learner $M$ is said to 
\Vacf\ learn $\mathcal{C}$ iff $M$
outputs on any text $T$ for every
$L \in \mathcal{C}$ only finitely
many indices, and for almost all
$n$, $W_{M(T[n+1])}$ is a finite
variant of $L$. 
\end{defn}

\begin{exmp}
Case and Smith \cite{Cas83} showed that
\Vacf\ and \Exf\ learning of recursive
functions are equivalent.  However, this 
equivalence does not extend to all classes 
of r.e.\ sets.  Take, for example,
the class $\mathcal{C} = \{\{e\} \oplus \natnum: e\in\natnum\}
\cup \{\{e\} \oplus \{x: x \leq |W_e|\}:
e\in\natnum\}$.  $\mathcal{C}$ is \Vac\
learnable: on any input $\sigma$ whose
range is of the form $\{e\} \oplus D$, determine 
whether $\max(D) > |W_{e,|\sigma|}|$;
if so, conjecture $\{e\} \oplus \natnum$;
otherwise, conjecture $\{e\} \oplus \{x: x \leq |W_e|\}$.
If $range(\sigma)$ does not contain any
even number, conjecture $range(\sigma)$.

On the other hand, $\mathcal{C}$ is not
\Exf\ learnable.  Assume by way of a
contradiction that a recursive learner $M$
\Exf\ learns $\mathcal{C}$.  Using
$K$ as an oracle, one can 
determine for any $e$ whether $W_e$
is finite.  By the assumption that
$M$ is an \Exf\ learner, one can
enumerate a text $T$ for $L_e = \{e\} 
\oplus \{x: x \leq |W_e|\}$
until at least one of the following holds.
\begin{enumerate}
\item There is some $m$ such that for all
$x > m$, $x \notin W_e$.  This immediately
implies that $W_e$ is finite.   
\item For some $\sigma \in (L_e \cup \{\#\})^*$
such that $\sigma$ is a prefix of $T$,
it holds that for all $\eta \in (L_e \cup \{\#\})^*$,
$M(\sigma\eta) = M(\sigma)$; in other
words, $\sigma$ is a locking sequence
for $L_e$. 
\end{enumerate}
Now one can use $K$ again to determine
whether or not there exists an $\eta\in(\{e\}\oplus\natnum)^*$
such that $M(\sigma\eta) \neq M(\sigma)$.
Suppose that $|W_e|$ is finite.  Then
$\{e\} \oplus \natnum$ is not a finite
variant of $L_e$; furthermore, as 
$M$ must \Exf\ learn $\{e\} \oplus \natnum$,
there must exist some $\eta \in (\{e\} \oplus \natnum)^*$
for which $M(\sigma\eta) \neq M(\sigma)$.
Suppose, on the other hand, that $|W_e|$
is infinite.  Then $L_e = \{e\} \oplus \natnum$,
so that by the locking sequence property of
$\sigma$, $M(\sigma\eta) = M(\sigma)$
for all $\eta \in (\{e\} \oplus \natnum)^*$.  
Hence the \Exf\ learnability of $\mathcal{C}$ would
imply that $\{e: |W_e| < \infty\}$ is
Turing reducible to $K$, which
is known to be false \cite{Rog}.~\qed
\end{exmp}

\begin{thm}\label{bcfinfinite}
Suppose there is a recursive learner that \BCf\
learns $\mathcal{C}$ and outputs on
every text for any $L \in \mathcal{C}$
at least one index infinitely often.
Then $\mathcal{C}$ is \BCfPart\
learnable.
\end{thm}

\proof
Let $M$ be a recursive \BCf\ learner of $\mathcal{C}$
such that $M$ outputs on every text
for any $L \in \mathcal{C}$ at least
one index infinitely often.  Define a
learner $N_1$ as follows.

On any given text $T$ for some $L \in \mathcal{C}$, 
let $e_n = M(T[n+1])$.  Let $F_0,F_1,F_2,
\ldots$ be a one-one enumeration of all finite sets.  
On input $T[k+1]$, $N_1$ outputs a canonical index 
$d_{e_k,l}$ for $W_{e_k} \cup F_l$
(respectively $g_{e_k,l}$ for $W_{e_k} - F_l$)
at least $m$ times iff the following conditions
hold:
\begin{enumerate}
\item $M$ outputs $e_k$ at least $l+1$ times;
\item there is a stage $s > m$ such that for all 
$x < m$, $x \in range(T[s+1])$
iff $x \in W_{e_k,s} \cup F_l$ (respectively
$x \in range(T[s+1])$ iff $x \in W_{e_k,s} - F_l$).   
\end{enumerate}
It will be shown that $N_1$ has the following
two learning properties: first, it $BC^*$ learns 
$\mathcal{C}$; second, it outputs at least
one correct index infinitely often; third,
it outputs an incorrect index only finitely
often.  Consider any $e_k$.

First, suppose that $W_{e_k}$ is not a finite
variant of $L$.  Then $M$ outputs $e_k$ only
finitely often.  Further, $N_1$ will consider
sets of the form $W_{e_k} \cup F_l$ or
$W_{e_k} - F_l$ for only finitely many
$F_l$.  Since, for each such $W_{e_k} \cup F_l$
(or $W_{e_k} - F_l$), item 2.\ will be satisfied  
for only finitely many $m$, it follows that
$N_1$ will conjecture a set of the form $W_{e_k} 
\cup F_l$ or $W_{e_k} - F_l$ only finitely often.

Second, suppose that $W_{e_k}$ is a finite
variant of $L$.  Then for any $F_l$, 
$W_{e_k} \cup F_l$ and $W_{e_k} - F_l$ are both 
finite variants of $L$.  Hence $N_1$ preserves its \BCf\
learning property by outputting any indices
for $W_{e_k} \cup F_l$ or $W_{e_k} - F_l$. 
Moreover, $M$ outputs infinitely often at least 
one index $e_h$ such that $W_{e_h}$ is a finite 
variant of $L$.  If $L = W_{e_h} \cup F_c$ 
(respectively $L = W_{e_h} - F_c$)
for some $F_c$, then $N$ will consider
$W_{e_h} \cup F_c$ (respectively
$W_{e_h} - F_c$) after $M$ has output
$e_h$ at least $c+1$ times.  As
$W_{e_h} \cup F_c$ (respectively $W_{e_h} - F_c$)
satisfies item 2.\ for almost all $m$,
$N_1$ will output at least one index for
$L$ infinitely often.

Third, suppose that for some $F_l$, neither 
$W_{e_k} \cup F_l$ nor $W_{e_k} - F_l$ is equal 
to $L$.  Then $W_{e_k} \cup F_l$ and
$W_{e_k} - F_l$ will satisfy Condition 2.\ 
for all but finitely many $m$, and so
$N_1$ will output a canonical index for
$W_{e_k} \cup F_l$ or $W_{e_k} - F_l$
only finitely often.  This establishes
the three learning properties of $N_1$.

Using a padding technique, one can define a 
further learner $N$ such that
$N$ preserves the \BCf\ learning property 
of $N_1$; further, if $e'_h$ is the 
minimum index that $N_1$ outputs infinitely 
often on $T$, then there is a $h'$ with
$e'_h = e_{h'}$ such that $N$ will output 
$pad(e'_{h'}, d_{h'})$ infinitely often, 
and every other index is output
only finitely often.  Therefore
$N$ is both a \BCf\ and a \Part\
learner of $\mathcal{C}$.~\qed

\begin{cor} \label{vacstarimpbcpart}
If a class $\mathcal{C}$ of r.e.\ sets is \Vacf\
learnable, then $\mathcal{C}$ is \BCfPart\
learnable. 
\end{cor}

\begin{exmp}
Case and Smith \cite{Cas83} showed that
the class of recursive functions
$\mathcal{F} = \{f: f\ \mbox{is recursive}
\wedge \forall^{\infty}x[f = \varphi_{f(x)}]\}$ is \BC\ learnable
but not \Exf\ learnable.  By the equivalence
of \Exf\ and \Vacf\ in the setting of learning 
recursive functions, $\mathcal{F}$ is also not \Vacf\
learnable.  Furthermore, by Theorem \ref{bcfinfinite},
the class $\mathcal{F}$ witnesses the separation of
\Vacf\ and \BCfPart\ learnability.     
\end{exmp}

\medskip
\noindent
The following proposition shows that
two relatively strong learning criteria
can be synthesized to produce quite a
strict learning criterion. 

\begin{prop}\label{vacfwpart}
If a class $\mathcal{C}$ of r.e.\ sets is 
\VacfWPart\ learnable, then $\mathcal{C}$ 
is \Vac\ learnable.
\end{prop}

\proof
Assume that $M$ is a \VacfWPart\ learner of
$\mathcal{C}$.  Define a new learner $N$
as follows.  On input $\sigma$, let
$e_0,e_1,\ldots,e_k$ be all the distinct
conjectures of $M$ on prefixes of
$\sigma$.  For each $e_i$, let
$p_i$ be the maximum number
such that for all $x < p_i$, $x \in W_{e_i,|\sigma|}$
holds iff $x$ is contained in $range(\sigma)$. 
Furthermore, let $q = \max(\{p_i: 0 \leq i \leq k\})$
and $m$ be the least index such that
$p_m = q$; $N$ then outputs $e_m$.

Let $d_0,\ldots,d_l$ be all the distinct
conjectures of $M$ on some text $T$
for an $L \in \mathcal{C}$.  Since $M$
is a \WPart\ learner, it must output
at least one index for $L$.
on $T$.  Consider any $d_i,d_j$ such
that $W_{d_i} \neq L$ and $W_{d_j} = L$.
Let $z_i$ be the maximum number such
that for all $x < z_i$, $x \in W_{d_i}$ holds
iff $x \in L$.  Then on almost all text prefixes
$T[s]$, there must exist some $y_j > z_i$ 
such that for all $x < y_j$, $x \in W_{d_j, s+1}$
iff $x$ is contained in $range(T[s])$.
As there are only finitely many incorrect
indices that $M$ outputs, it follows that
$N$ will almost always output some index
$d_c$ for which $W_{d_c} = L$.  Therefore
$N$ is a \Vac\ learner of $\mathcal{C}$.~\qed    

\medskip
\noindent
The following proposition implies that
vacillatory learning cannot in general
be combined with partial learning;
in other words, a vacillatorily learnable
class may not necessarily be
vacillatorily as well as partially
learnable at the same time.

\begin{prop}\label{vacfpart}
If a class $\mathcal{C}$ of r.e.\ sets is 
\VacfPart\ learnable, then $\mathcal{C}$ 
is \Ex\ learnable.
\end{prop}

\proof
If $M$ is a recursive learner of $\mathcal{C}$
such that on any text $T$ for some $L \in \mathcal{C}$, 
$M$ outputs only finitely many indices and
outputs exactly one index $d$ for $L$ infinitely often, then 
$M$ almost always outputs $d$ on $T$.~\qed

\begin{exmp}\label{sepvacfpart}
The class of all cofinite sets is \Exf\ learnable (and
hence \Vacf\ learnable) but it is not \Ex\ learnable.
By Prop \ref{vacfpart}, this class is also not \VacfPart\ learnable.
\end{exmp}

\section{Conclusion}

This paper studied conditions under which various forms of partial
learning can be combined with models of approximation and with \BCf\ learning. 
For learning of recursive functions, it positively resolved Fulk and
Jain's open question on whether the class of all recursive functions
can be approximately learnt and \BCf\ learnt at the same time. For
learning r.e.\ languages, three notions of approximate learning were
introduced and studied.
However, questions on the combinability 
of some pairs of learning constraints remain open.
In particular, it is unknown whether or not 
every \BCf\ learnable class of r.e.\ languages has a learner that
is both \BCf\ and \Part.

\end{document}